\pdfoutput=1
\documentclass[12pt]{article}
\usepackage[numbers]{natbib} 
\usepackage{graphicx}
\usepackage{url} 
\usepackage{caption} 
\usepackage{caption}
\usepackage{subcaption}
\usepackage[dvipsnames]{xcolor}
\usepackage{soul}
\usepackage[linesnumbered,ruled]{algorithm2e}
\usepackage{amsfonts}
\usepackage{amsmath}
\usepackage{wrapfig}
\usepackage{mathtools}

\usepackage{enumitem}
\usepackage{soul}
\newtheorem{prop}{Proposition}

\newcommand{\blind}{0}

\begin{document}

\def\spacingset#1{\renewcommand{\baselinestretch}%
{#1}\small\normalsize} \spacingset{1}


\if0\blind
{
  \title{\bf A Federated Data Fusion-Based Prognostic Model for Applications with Multi-Stream Incomplete Signals} 
  \author{Madi Arabi\\
  Edward P. Fitts Department of Industrial and Systems Engineering, \\North Carolina State University, Raleigh, U.S. \\
    and \\
    Xiaolei Fang \\
    Edward P. Fitts Department of Industrial and Systems Engineering, \\North Carolina State University, Raleigh, U.S. }
  \date{}
  \maketitle
} \fi

\if1\blind
{
  \bigskip
  \bigskip
  \bigskip
  \begin{center}
    {\LARGE\bf Title}
\end{center}
  \medskip
} \fi

\bigskip
\begin{abstract}

Most prognostic methods require a decent amount of data for model training. In reality, however, the amount of historical data owned by a single organization might be small or not large enough to train a reliable prognostic model. To address this challenge, this article proposes a federated prognostic model that allows multiple users to jointly construct a failure time prediction model using their multi-stream, high-dimensional, and incomplete data while keeping each user's data local and confidential. The prognostic model first employs multivariate functional principal component analysis to fuse the multi-stream degradation signals. Then, the fused features coupled with the times-to-failure are utilized to build a (log)-location-scale regression model for failure prediction. To estimate parameters using distributed datasets and keep the data privacy of all participants, we propose a new federated algorithm for feature extraction. Numerical studies indicate that the performance of the proposed model is the same as that of classic non-federated prognostic models and is better than that of the models constructed by each user itself.

\end{abstract}

\noindent%
{\it Keywords:}  Federated Learning, Federated Data Fusion, Remaining Useful Life 
\vfill

\newpage
\spacingset{1.3} 

\section{Introduction}

Industrial prognostic aims to predict the failure time of machines by utilizing their degradation signals. This is typically achieved by establishing a statistical learning model that maps the degradation signals of machines to their time-to-failure (TTFs) \cite{gebraeel2004residual,sikorska2011prognostic,zhou2023federated}. Similar to that of many other statistical learning models, the implementation of prognostic models usually consists of two steps: model training and real-time monitoring (also known as model testing or deployment). Model training focuses on using a historical dataset that comprises the degradation signals and TTFs of some failed machines to estimate the parameters of the prognostic model; real-time monitoring feeds the real-time degradation signals from a partially degraded onsite machine into the prognostic model trained earlier to predict its TTF or TTF distribution. Most existing prognostic models assume that a historical dataset from a decent number of failed machines is available for model training \cite{9032345, gebraeel2008neural,4781602,fang2017scalable,fang2017multistream,zhou2023supervised}. In reality, however, the amount of historical data owned by a single organization (e.g., a company, a university lab, a factory, etc.) might be small or not large enough to train a reliable prognostic model,which is typically known as the \textit{limited data availability} challenge. To address this challenge, it is beneficial to investigate the possibility of allowing multiple geographically distributed organizations/users/participants to collaboratively train a prognostic model using their data together.

A straightforward solution that enables multiple geographically distributed users to jointly train a prognostic model is cloud computing-based prognostic, which works as follows. Specifically, all the users first upload their data to a server on the cloud. Next, the server aggregates the data from all users and uses the aggregated data to train a prognostic model on the cloud. Finally, the trained model is shared with all users for real-time monitoring. Unfortunately, cloud computing-based prognostic is usually not feasible in real-world applications since it cannot protect the data privacy of participants. This is because many companies have concerns over sharing their data, and some are not even allowed to upload their data to the cloud due to various privacy protection policies and regulations \cite{pardau2018california,voigt2017eu,chik2013singapore}. One solution to address this challenge is \textit{Federated Learning}, which allows multiple organizations to use their isolated data to jointly train machine learning models while keeping each participant's data local and confidential \cite{kontar2021internet, konevcny2016federated,konevcny2016federated2,mcmahan2016federated,li2019survey,yang2019federated,li2020review,li2020federated,briggs2021review}. However, integrating federated learning into prognostic is nontrivial since it requires the development of new privacy-preserving statistical learning models or parameter estimation algorithms under the distributed computing framework. 

In the recent literature, there have been some efforts in developing federated learning-based prognostics \cite{bagheri2020unified, 10091163,9343286,10159214,9904020,chen2023bearing,guo2022fedrul,su2022}. However, many of these models focus on modeling single-stream degradation signals \cite{10091163,9343286,10159214,9904020}, which means they cannot be applied to applications with degradation signals from multiple sensors. As discussed and verified in many articles \cite{liu2013data, fang2017multistream,fang2017scalable}, prognostic models based on multi-stream degradation signals usually achieve a better performance than those work on single-stream signals since multi-stream (multi-sensor) data may capture different aspects of the underlying degradation process. Also, advances in sensing technology have enabled the industry to monitor their equipment using multiple sensors, which generate multi-stream degradation signals. Some other works focus on developing prognostic models based on multi-stream degradation signals \cite{bagheri2020unified,chen2023bearing,guo2022fedrul,su2022}. However, many of them fail to fuse the information from multi-stream data, which might compromise the performance of their prognostic models. As discussed in articles \cite{fang2017multistream, liu2013data}, data fusion or dimension reduction is usually necessary for the prognostic of machines with multi-stream degradation signals. This is because degradation signals are typically high-dimensional, which implies the number of parameters to be estimated in prognostic models is large and often far more than the size of samples in the historical training data (i.e., the small $n$ large $p$ issue in statistical learning). Also, high-dimensional degradation signals usually contain redundant/overlapped information that provides the possibility for data fusion to reduce the number of features used to construct prognostic models. The redundant information is manifested in terms of correlations within and among degradation signals: the observations of a single degradation signal are usually auto-correlated and observations among different degradation signals are often cross-correlated. A few papers have developed federated data fusion algorithms before constructing federated prognostic models. For example, the authors in \cite{su2022} proposed a federated feature extraction algorithm based on randomized singular value decomposition. However, this article assumes that degradation signals are complete, which implies it cannot be applied to applications whose degradation signals have missing observations. In real-world applications, incompleteness is a common characteristic of degradation signals. One of the reasons is that many complex assets operate in harsh environments that often have a significant impact on the quality of the raw data, due to errors in data acquisition, communication, read/write operations, etc, which results in degradation signals with significant levels of missing and corrupt observations. Another reason for data incompleteness is the failure time truncation. Specifically, the failure times of machines vary from one to another, and no degradation signal can be observed beyond the failure times. As a result, the lengths of degradation signals of different machines are usually different. Unfortunately, many prognostic models require the lengths of degradation signals to be the same. To the best of our knowledge, few works have focused on developing prognostic models that work with incomplete degradation signals. To address the aforementioned challenges, this paper proposes a federated prognostic model that can fuse multi-stream incomplete degradation signals and use the fused features for TTF distribution prediction. 

The proposed federated prognostic model comprises two sequential steps: data fusion and prognostic model construction. Data fusion focuses on fusing multi-stream high-dimensional degradation signals and provides low-dimensional features, while prognostic model construction is accomplished by mapping TTFs to the features extracted earlier using a statistical learning model. In this article, we use multivariate functional principal component analysis (MFPCA) for data fusion. MFPCA is a nonparametric dimension reduction method that captures the joint variation of multi-stream functional data (multi-stream degradation signals in this paper) and provides low-dimensional features known as MFPC-scores \cite{fang2017scalable}. It can capture the auto-correlation within each degradation signal as well as cross-correlation among different degradation signal streams. MFPCA provides low-dimensional fused features known as MFPC-scores, which are then used for prognostic model construction. In this article, we use (log)-location-scale (LLS) regression to build the prognostic model, in which the response variable is the TTF and the predictors are MFPC-scores. LLS regression has been widely used in reliability engineering and survival analysis since their response variables are general and include a variety of TTF distributions, such as (log)-normal, (log)-logistics, Smallest Extreme Value (SEV), and Weibull \cite{doray1994ibnr}. Prognostic models based on MFPCA and LLS regression have been discussed and their effectiveness has been extensively validated in existing articles \cite{fang2017scalable,fang2017multistream,fang2015adaptive,fang2021multi}. However, all of these models use a centralized dataset for model training, and none of them can be used under federated learning settings. Our contribution in this paper is to propose a federated algorithm that enables multiple users to use their (multi-stream) incomplete data to collaboratively conduct MFPCA and extract fused features (i.e., MFPC-scores) while keeping every user's data local and confidential. 

The proposed federated feature extraction algorithm is inspired by the classic singular value decomposition (SVD)-based MFPC-scores computation method. The classic SVD-based method works with a degradation signal matrix that aggregates signals from all failed machines in the training dataset. First, it removes the mean of signals, which is known as data centralization. Next, it conducts SVD on the centered signal matrix to get singular values and singular vectors. Third, MFPC-scores are computed by multiplying the centered signal matrix by singular vectors, and the eigenvalues are often used to determine the number of dominant principal components to be kept. Apparently, the classic SVD-based method cannot be applied in federated learning settings since it requires data from all users to be aggregated first. Also, it assumes the aggregated signal matrix is complete since the centralization and multiplication cannot be done if signals have missing entries. Thus, a new federated algorithm that can work with incomplete degradation signals needs to be developed. From the perspective of linear algebra and geometry, the singular vectors of the classic SVD-based method span the low-dimensional dominant subspace in the high-dimensional space spanned by the centered degradation signals. Also, the singular vectors are a special set of orthonormal bases of the dominant subspace (the dominant subspace has numerous sets of basis vectors, and the singular vectors are one of these sets). Inspired by this, we first propose a federated dominant subspace detection algorithm that can use uncentered and incomplete degradation signals to extract a general set of basis vectors of the dominant subspace. This is achieved by developing a new federated incremental SVD algorithm, which is an iterative method that allows each user to update basis vectors locally using its own data. Since only the basis vectors are shared among different users, data privacy can be protected. With the general set of basis vectors provided by the federated dominant subspace detection algorithm, we then propose a federated algorithm for MFPC-scores computation.

The rest of the paper is organized as follows. In section \ref{sec: prognostic}, we introduce the MFPCA- and LLS regression-based prognostic model. Section \ref{sec: para} discusses the feature extraction method, in which Section \ref{sec: sub: para_center} summarizes the computation of MFPC-scores with an aggregated dataset while Section \ref{sec: sub: para_decenter} presents our proposed federated feature extraction method. In Sections \ref{sec:sim} and \ref{sec:case}, the performance of the proposed federated prognostic model is evaluated using a simulated dataset and a degradation dataset from the NASA data repository. Finally, Section \ref{sec:conc} concludes. 

\section{The Prognostic Model and Parameter Estimation}\label{sec: prognostic}


This article proposes a federated prognostic model that can be jointly constructed by multiple geographically distributed users using their incomplete degradation signals while keeping each user's data local and confidential. Similar to some existing articles \cite{fang2017multistream,fang2021multi}, the prognostic model used in this paper is based on functional (log)-location-scale regression \cite{meeker1998statistical}, which regresses a system's TTF against its multi-stream degradation signals. To improve the performance of the prognostic model and simplify its parameter estimation, we transform functional LLS regression into classic LLS regression. This is achieved by first employing Multivariate Functional Principal Component Analysis (MFPCA) to fuse the (multi-stream) high-dimensional degradation signals and provide low-dimensional features; then, the TTF of each system is regressed against its features from MFPCA using regular LLS regression. Articles \cite{fang2017multistream, fang2021multi} have proved that such a transformation neither results in the loss of information nor compromises the performance of the prognostic model.

 We assume there are $I$ users jointly constructing the prognostic model, and user $i$ has the historical degradation signals and TTFs of $J_i$ failed systems/assets/components. Also, we assume that the failed systems of all users are identical, and each of them is monitored by $P$ sensors, which generate $P$-channel degradation signals (without loss of generality, we assume that each sensor generates one channel degradation signal). Let ${x}_{ijp}(t)$ be the degradation signal from sensor $p$ of system $j$ of user $i$, where $i=1,\ldots, I$, $j=1,\ldots, J_i$, and $p=1,\ldots, P$, $t \in [0,T]$. Also, let $\tilde{y}_{ij}$ be the TTF of the system $j$ from user $i$. To map TTF to the degradation signals, we employ the following functional LLS regression model:
\begin{equation}\label{eq:flls}
    {y}_{ij} = \gamma_0 +\int_0^T \boldsymbol{\gamma}(t)^\top \boldsymbol{x}_{ij}(t)dt +\sigma\epsilon_{ij},
\end{equation}

\noindent where ${y}_{ij}=\tilde{y}_{ij}$ if the TTF follows a location-scale distribution and ${y}_{ij}=\log(\tilde{y}_{ij})$ if it follows a (log)-location-scale distribution. $\gamma_0$ is the intercept, $\boldsymbol{\gamma}(t)=(\gamma_1(t), \gamma_2(2),...,\gamma_P(t))^\top$ is the regression coefficient function, and
$\boldsymbol{x}_{ij}(t)=(x_{ij1}(t),x_{ij2}(t),...,x_{ijP}(t))^\top$ is the concatenated degradation signals from all $P$ sensors of system $j$ of user $i$. 
$\sigma$ is the scale parameter, and $\epsilon_{ij}$ is
the random noise term with a standard location-scale density
$f(\epsilon)$. For example $f(\epsilon)=1/\sqrt{2\pi}\exp(-\epsilon^2)$ for normal distribution and $f(\epsilon)=\exp(\epsilon-\exp(\epsilon))$ for SEV distribution. 

Based on the Kosambi-Karhunen-Lo\`eve theorem \cite{karhunen1947ueber}, the concatenated degradation signal $\boldsymbol{x}_{ij}(t)$ can be decomposed as follows:
\begin{equation}\label{eq3}
    \boldsymbol{x}_{ij}(t) = \boldsymbol{\mu}(t) + \sum_{k=1}^\infty \boldsymbol{\zeta}_{ijk}\boldsymbol{\Psi}_k(t),  \quad t \in [0,T]
\end{equation}

\noindent where $\boldsymbol{\mu}(t)=(\mu_1(t), \mu_2(t),...,\mu_P(t))^\top$ is the mean function; ${\zeta}_{ijk}=\int_0^T (\boldsymbol{x}_{ij}(t)-\boldsymbol{\mu}(t))^\top\boldsymbol{\Psi}_k(t)dt$ is the feature (known as MFPC-score). $\boldsymbol{\Psi}_k(t) = (\Psi_{k1}(t),...,\Psi_{kP}(t))$ is the $k$th eigen function of a $P \times P$ block covariance matrix    $\boldsymbol{C}(t,t')=
\begin{bmatrix}
    C_{1,1}(t,t')& ...& C_{1,P}(t,t')\\
    \vdots & \ddots &\vdots\\
    C_{P,1}(t,t')&...& C_{P,P}(t,t')\\
\end{bmatrix}$, where the $(g,h)$th entry ${C}_{g,h}(t,t')$ is the covariance function between sensor $g$ and $h$, $g=1,\ldots,P, h=1,\ldots,P$. Usually, it is sufficient to keep the first $K$ eigenfunctions corresponding to the $K$ largest eigenvalues. As a result, Equation \eqref{eq3} can be expressed as follows: $
 \boldsymbol{x}_{ij}(t) \approx \boldsymbol{\mu}(t) + \sum_{k=1}^K \boldsymbol{\zeta}_{ijk}\boldsymbol{\Psi}_k(t)$, where $K$ can be determined using Fraction-of-Variance Explained (FVE) \cite{fang2015adaptive} or cross-validation \cite{fang2017multistream}. Since the set of eigenfunctions $\{\boldsymbol{\Psi}_k(t)\}_{k=1}^\infty$ forms a complete set of orthonormal basis functions, the unknown regression coefficient function $\boldsymbol{\gamma}(t)$ can be decomposed as $\boldsymbol{\gamma}(t)=\sum_{k=1}^\infty\beta_k\boldsymbol{\Psi}_k(t)\approx\sum_{k=1}^K\beta_k\boldsymbol{\Psi}_k(t)$ \cite{yao2005functional}. As a result, the function LLS regression model in Equation \eqref{eq:flls} can be simplified to the following classic LLS regression (the proof is provided in the appendix, which can also be found in \cite{fang2017multistream}):
\begin{equation}\label{eq5}
y_{ij} = \beta_0 +\boldsymbol{\beta}^\top\boldsymbol{\zeta}_{ij} +\sigma\epsilon_{ij},
\end{equation}
\noindent where $\beta_0$ is the intercept, $\boldsymbol{\beta}=(\beta_1, \beta_2,...,\beta_K)^\top \in \mathbb{R}^K$ is the coefficient vector, and $\boldsymbol{\zeta}_{ij}=(\zeta_{ij1},...,\zeta_{ijK})^\top \in \mathbb{R}^K$ are the MFPC-scores of the $j$th system of user $i$.

The parameter estimation of the prognostic model consists of two steps: \textit{(i) feature extraction} and \textit{(ii) regression parameters estimation}. The first step, feature extraction, focuses on using the decentralized historical degradation signals from multiple users (i.e., $\boldsymbol{x}_{ij}(t)$) to calculate the MFPC-scores (i.e., $\boldsymbol{\zeta}_{ij}$ in Equation \eqref{eq5}). The second step, regression parameters estimation, uses the features from the first step (i.e., $\boldsymbol{\zeta}_{ij}$) and the historical TTFs (i.e., $y_{ij}$  in Equation \eqref{eq5}) to estimate the regression parameters (i.e., $\beta_0,\boldsymbol{\beta}, \sigma$). Since the historical data is decentralized, both the feature extraction and coefficient estimation algorithms should be federated, which means that they can use data from all the users while keeping each user's data local and confidential. In this article, our contribution is to propose a federated feature extraction algorithm for applications with incomplete degradation signals, which will be discussed in Section \ref{sec: para}. Regarding the second step, regression parameters estimation, we will employ the federated algorithm developed in \cite{su2022}.

\section{The Federated Feature Extraction Algorithm}\label{sec: para}

In this section, we propose a federated feature extraction algorithm that computes the MFPC-scores (i.e., $\boldsymbol{\zeta}_{ij}$ in Equation \eqref{eq5}) using the decentralized incomplete degradation signals from multiple users while keeping each user's data local and confidential. To help understand the proposed federated algorithm, we will first review how to extract the MFPC-scores using centralized complete data (i.e., the data from all users are merged together for feature extraction) in Section \ref{sec: sub: para_center}. After that, the proposed federated algorithm using decentralized data will be discussed in Section \ref{sec: sub: para_decenter}.

\subsection{Feature Extraction Using Centralized Data}\label{sec: sub: para_center}

Recall that the degradation signal from senor $p$ of the system $j$ of user $i$ is denoted as ${x}_{ijp}(t), t\in[0,T]$. Let vector $\boldsymbol{x}_{ijp}=({x}_{ijp}(\tau_1^{(p)}),{x}_{ijp}(\tau_2^{(p)}),\ldots,{x}_{ijp}(\tau_{n_p}^{(p)}))^\top\in\mathbb{R}^{n_p}$ be its complete discrete observations, where $0\leq\tau_1^{(p)}\leq\tau_2^{(p)}\leq\ldots\leq\tau_{n_p}^{(p)}\leq T$ are the discrete observation time stamps, $n_p$ is the number of observations for sensor $p$, $p=1,\ldots, P$, $i=1,\ldots, I$, and $j=1,\ldots, J_i$. Then, the concatenated degradation signal observations from all $P$ sensors of system $j$ belonging to user $i$ can be denoted as $\boldsymbol{x}_{ij}=(\boldsymbol{x}_{ij1}^\top, \boldsymbol{x}_{ij2}^\top,\ldots, \boldsymbol{x}_{ijP}^\top)^\top\in\mathbb{R}^{\sum_{p=1}^Pn_p}$. Furthermore, the degradation signal matrix after combing the data from all users can be expressed as $\boldsymbol{X}=(\boldsymbol{x}_{11},\boldsymbol{x}_{12},\ldots, \boldsymbol{x}_{1J_1},\boldsymbol{x}_{21},\boldsymbol{x}_{22},\ldots, \boldsymbol{x}_{2J_2},\ldots,\boldsymbol{x}_{I1},\boldsymbol{x}_{I2},\ldots, \boldsymbol{x}_{IJ_I})\in\mathbb{R}^{N\times J}$, where $N=\sum_{p=1}^Pn_p$ is the total number of observations from all $P$ sensors, and $J=\sum_{i=1}^IJ_i$ is the total number of systems from all $I$ users.  


Given the centralized signal matrix $\boldsymbol{X}$, the MFPC-scores can be computed following the steps below \cite{fang2017scalable,fang2017multistream,fang2015adaptive}: 

\begin{enumerate}[label=(\alph*),noitemsep]
    \item Compute the mean vector of the signal matrix, i.e., $\boldsymbol{\bar{x}}=\frac{1}{J}\sum_{i=1}^I\sum_{j=1}^{J_i}\boldsymbol{x}_{ij}\in\mathbb{R}^N$.

    \item Center the signal matrix by subtracting the mean vector from each column of the signal matrix, i.e., $\boldsymbol{\tilde{X}}=\boldsymbol{X}-\boldsymbol{\bar{X}}\in\mathbb{R}^{N\times J}$, where $\boldsymbol{\bar{X}}$ has $J$ columns, each of which is $\boldsymbol{\bar{x}}$.

     \item Compute the covariance matrix of the centered signal matrix $\boldsymbol{C}=\boldsymbol{\tilde{X}}\boldsymbol{\tilde{X}}^\top\in\mathbb{R}^{N\times N}$ and conduct eigen decomposition $\boldsymbol{C}=\boldsymbol{U}\boldsymbol{\Sigma}\boldsymbol{U}^\top$, where $\boldsymbol{U}\in\mathbb{R}^{N\times N}$ is a matrix whose columns are the eigenvectors, and $\boldsymbol{\Sigma}\in\mathbb{R}^{N\times N}$ is a diagonal matrix whose diagonal entries are the eigen values $\lambda_1,\lambda_2,\ldots,\lambda_N$ and  $\lambda_1\geq\lambda_2\geq\ldots\geq\lambda_{N}$. Another way to compute the eigenvectors and eigenvalues is to conduct SVD on $\boldsymbol{\tilde{X}}$. That is, $\boldsymbol{\tilde{X}}=\boldsymbol{U}\boldsymbol{S}\boldsymbol{V}$, where $\boldsymbol{U}\in\mathbb{R}^{N\times N}$ is a matrix whose columns are the left singular vectors (same as the eigenvectors from the eigendecomposition of $\boldsymbol{C}$), $\boldsymbol{S}\in\mathbb{R}^{N\times J}$ is a diagonal matrix whose entries are the singular values (i.e., the square root of eigenvalues $\sqrt{\lambda_1}\geq\sqrt{\lambda_2}\geq\ldots\geq\sqrt{\lambda_{N}}$), and $\boldsymbol{V}\in\mathbb{R}^{N\times J}$.

     \item Use the fraction-of-variance-explained \cite{fang2015adaptive} or cross-validation \cite{fang2017multistream} to select the number of dominant principal components (denoted as $K$, $K\leq\min\{N,J\}$) and keep the first $K$ eigenvectors -- that is -- $\boldsymbol{U}_{dominant}=\boldsymbol{U}_{[1:N,1:K]}\in\mathbb{R}^{N\times K}$. Here, the subscript $_{[1:N,1:K]}$ represents the first $K$ columns.

     \item Calculate the MFPC-scores using $\boldsymbol{Z}=\boldsymbol{U}_{dominant}^\top\boldsymbol{\tilde{X}}\in\mathbb{R}^{K\times J}$. 

     
\end{enumerate}

Steps (a)-(e) summarize the process of computing MFPC-scores. Here, we give some remarks on the dominant eigenvectors $\boldsymbol{U}_{dominant}$. First, the dominant eigenvectors $\boldsymbol{U}_{dominant}\in\mathbb{R}^{N\times K}$ span a special $K$-dimensional subspace in $\mathbb{R}^{N}$. Here, ``special" refers to that, among all the $K$-dimensional subspaces in $\mathbb{R}^{N}$, the one spanned by $\boldsymbol{U}_{dominant}$ is the best subspace with respect to preserving information of $\boldsymbol{X}$. Specifically, among the numerous $K$-dimensional subspaces in $\mathbb{R}^{N}$, the one spanned by $\boldsymbol{U}_{dominant}$ preserves most information of $\boldsymbol{X}$ after projecting $\boldsymbol{X}$ onto it, i.e., $\boldsymbol{U}_{dominant}=\arg\min_{\boldsymbol{U}\in\mathbb{R}^{N\times K}}\|\boldsymbol{X}-\boldsymbol{U}\boldsymbol{U}^\top\boldsymbol{X}\|_F^2 \;\;s.t.\;\; \boldsymbol{U}^\top\boldsymbol{U}=\boldsymbol{I}$, where $\|\cdot\|_F$ is the Frobenius norm. For the sake of brevity, we refer to the $K$-dimensional subspace spanned by $\boldsymbol{U}_{dominant}$ as \textit{dominant subspace} hereafter. Second, the dominant subspace has numerous sets of basis vectors since any $K$ linearly independent vectors in the dominant subspace can span it, and $\boldsymbol{U}_{dominant}$ is one of these sets of basis vectors.

\subsection{Federated Feature Extraction Using Decentralized Data}\label{sec: sub: para_decenter}

\begin{figure}
    \centering
    \includegraphics[width=0.8\textwidth]{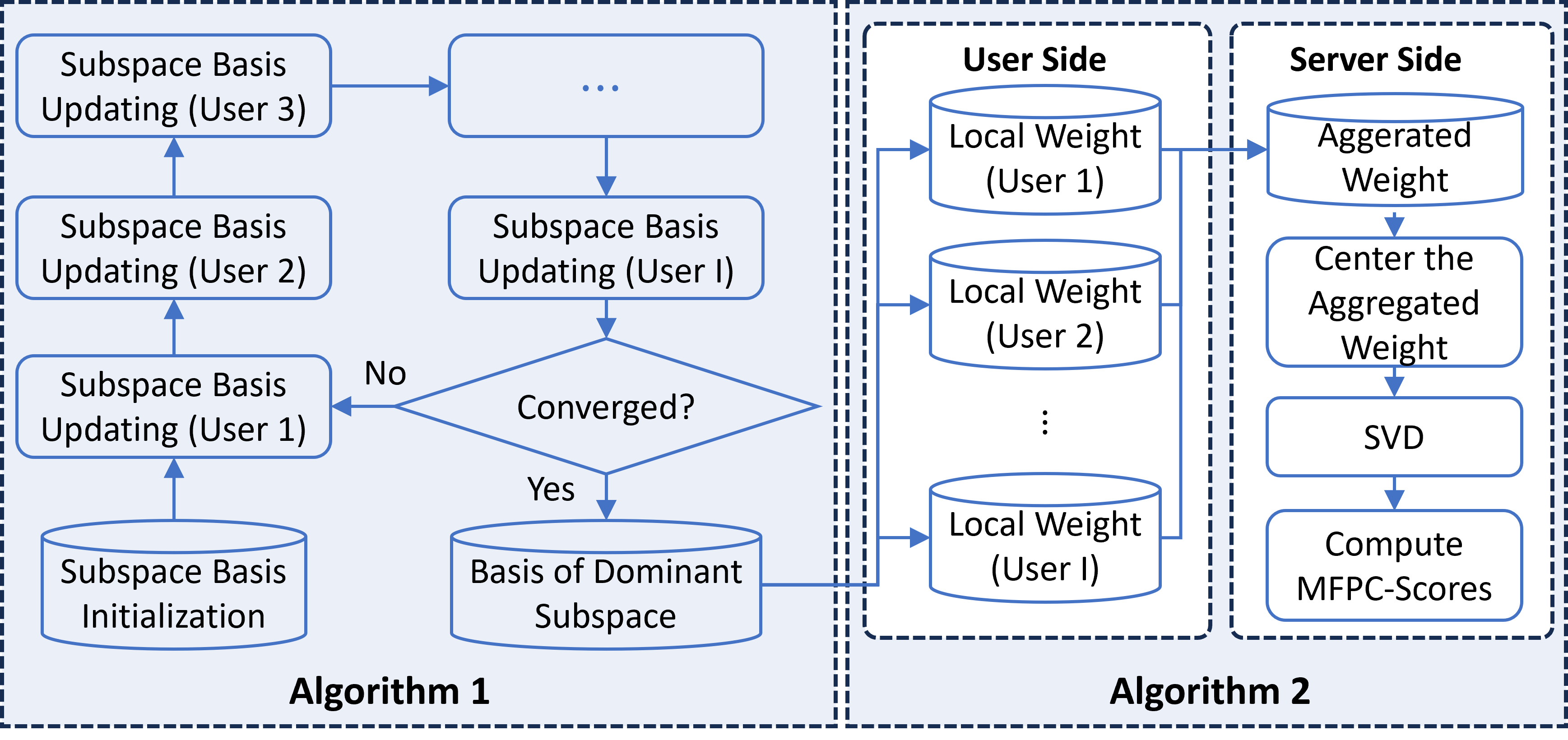}
    \caption{ Overview of proposed federated feature extraction method}
    \label{fig:framework}
\end{figure}

In this section, we propose a federated feature extraction method that enables multiple users to utilize their incomplete degradation signals to compute their MFPC-scores (i.e., $\boldsymbol{\zeta}_{ij}$  in Equation \eqref{eq5}) while keeping each user's data local and confidential. Inspired by the feature extraction method and the remarks discussed in Section \ref{sec: sub: para_center}, we first develop a federated algorithm to detect the dominant subspace of the degradation signal matrix $\boldsymbol{X}$ (Section \ref{sec:sub:sub:dominant}). This provides a general set of basis vectors of the dominant subspace. Next, we propose a federated algorithm that uses the general set of basis vectors of the dominant subspace and (incomplete and uncentered) degradation signals to calculate MFPC-scores (Section \ref{sec:sub:sub:mfpc}). Figure \ref{fig:framework} provides an overview of the proposed federated feature extraction algorithm.

\subsubsection{Federated Dominant Subspace Identification}\label{sec:sub:sub:dominant}

The federated dominant subspace identification algorithm was inspired by the incremental singular value decomposition (ISVD) method \cite{bunch1978updating, balzano2013grouse,brand2002incremental}. ISVD is an iterative algorithm that performs SVD on a matrix by successively using each row (or column) of the matrix to update the singular values and singular vectors. It first starts with a randomly initialized subspace. Next, the rows (or columns) of the matrix are projected onto the subspace one by one. After projecting a row (or column) onto the current subspace, the residual vector is analyzed. If no residual exists (i.e., the residual vector is a zero vector), it implies that the row (or column) vector is already within the current subspace. As a result, the current subspace does not need to be updated. However, if a residual does exist, it means that the current subspace needs to be revised. The revision is achieved by updating the current subspace so that after projecting the row (or column) vector onto the revised subspace, the norm of the new residual is minimized. The updating process is repeated (i.e., each row or column possibly gets involved more than once) until the norms of the residuals of all the rows (or columns) are zero or smaller than a predefined threshold. 

Following the notation definitions in Section \ref{sec: sub: para_center}, the concatenated degradation signal of user $i$'s $j$th system is denoted as $\boldsymbol{x}_{ij}\in\mathbb{R}^{N}$, where $i=1,\ldots, I$ and $j=1,\ldots, J_i$. Here, $I$ is the total number of users, and $J_i$ is the number of systems that user $i$ has. Due to data missing and failure truncation, the signal observations are incomplete. We use the set $\boldsymbol{\Omega}_{ij}$ and its complementary set $\boldsymbol{\Omega}_{ij}^C\subset\{1,2,\ldots, N\}\setminus \boldsymbol{\Omega}_{ij}$ to denote the indices of observations that are available and missing in $\boldsymbol{x}_{ij}$, respectively. The proposed federated dominant subspace identification algorithm works as follows. First,  we randomly initialize the subspace to be detected, which can be done by one of the users. Then, each user uses their own data to update the subspace sequentially until convergence. 

We take the signal from system $j$ of user $i$, $\boldsymbol{x}_{ij}$, as an example to explain how the subspace is updated. Let $\boldsymbol{U}_{old}\in\mathbb{R}^{N\times K}$ be the orthonormal matrix whose columns span the old subspace (before using $\boldsymbol{x}_{ij}$) and $\boldsymbol{U}_{new}\in\mathbb{R}^{N\times K}$ be the matrix spans the updated subspace (after using $\boldsymbol{x}_{ij}$). Here, $K$ is the dimension of the subspace, $0<K\leq N$, the value of which can be determined using cross-validation. We first project the incomplete signal $\boldsymbol{x}_{ij}$ onto the old subspace. This is achieved by identifying a linear combination of the basis vectors in $\boldsymbol{U}_{old}$ that can best approximate $\boldsymbol{x}_{ij}$:

\begin{equation}
    \hat{\boldsymbol{w}} = \arg\min_{\boldsymbol{w}} \|\boldsymbol{U}_{old}^{\boldsymbol{\Omega}_{ij}}\boldsymbol{w}-\boldsymbol{x}_{ij}^{\boldsymbol{\Omega}_{ij}}\|_{2}^{2},
\end{equation}

\noindent where $\hat{\boldsymbol{w}}\in\mathbb{R}^{K}$ is the weight vector, $\boldsymbol{x}_{ij}^{\boldsymbol{\Omega}_{ij}}\in\mathbb{R}^{|\boldsymbol{\Omega}_{ij}|}$ is the vector containing the observed entries of $\boldsymbol{x}_{ij}$, and $\boldsymbol{U}_{old}^{\boldsymbol{\Omega}_{ij}}\in\mathbb{R}^{|\boldsymbol{\Omega}_{ij}|\times K}$ is the matrix keeping the rows of $\boldsymbol{U}_{old}$ indexed by $\boldsymbol{\Omega}_{ij}$. As a result, the projected vector in the old subspace is $\boldsymbol{U}_{old}\hat{\boldsymbol{w}}$, and the missing entries of $\boldsymbol{x}_{ij}$ can be imputed as follows:

\begin{equation}
[\tilde{\boldsymbol{x}}_{ij}]_n =\begin{cases}
                    [{\boldsymbol{x}}_{ij}]_n, & n \in \boldsymbol{\Omega}_{ij}\\
                    [\boldsymbol{U}_{old}\hat{\boldsymbol{w}}]_n, & n \in \boldsymbol{\Omega}_{ij}^C
                    \end{cases}\\
\end{equation}

\noindent where $[\boldsymbol{a}]_n$ represents the $n$th entry of vector $\boldsymbol{a}$, and $\tilde{\boldsymbol{x}}_{ij}$ is the imputed signal of user $i$'s system $j$. Thus, the residual vector can be computed as follows:

\begin{equation}\label{eq:residual}
    {\boldsymbol{r}} = \tilde{\boldsymbol{x}}_{ij} - \boldsymbol{U}_{old}\hat{\boldsymbol{w}}.
\end{equation}

\noindent If the norm of the residual vector $\boldsymbol{r}\in\mathbb{R}^{N}$ (i.e., $\|\boldsymbol{r}\|\in\mathbb{R}$) is not zero, it means that vector $\boldsymbol{x}_{ij}$ is not fully within the subspace spanned by the columns of $\boldsymbol{U}_{old}$. Consequently, the subspace needs to be revised such that $\boldsymbol{x}_{ij}$ will be in the revised subspace or the residual after projecting $\boldsymbol{x}_{ij}$ onto the revised subspace will be minimized. The subspace revision can be achieved as follows. First, we construct a new matrix by appending $\tilde{\boldsymbol{x}}_{ij}$ to the end of $\boldsymbol{U}_{old}$ -- that is -- ${\boldsymbol{C}}=[\boldsymbol{U}_{old}\;\; \tilde{\boldsymbol{x}}_{ij}]\in\mathbb{R}^{N\times (K+1)}$. By doing so, we are sure that $\tilde{\boldsymbol{x}}_{ij}$ is within the $(K+1)$-dimensional subspace spanned by the columns of ${\boldsymbol{C}}$. However, since the dimension of the subspace we want is $K$, we will conduct SVD on ${\boldsymbol{C}}$ (i.e., ${\boldsymbol{C}}=\tilde{\boldsymbol{U}}\boldsymbol{\Sigma}\boldsymbol{V}$) and keep the first $K$ left singular vectors -- that is-- the updated subspace is spanned by the columns of $\boldsymbol{U}_{new}=\tilde{\boldsymbol{U}}{(1:N,1:K)}$, where $(1:N,1:K)$ represents all the $N$ rows and the first $K$ columns. 

One of the limitations of the above subspace updating method is that it is computationally intensive since SVD has to be conducted on an $N\times (K+1)$-dimensional matrix ${\boldsymbol{C}}$ for each update. Typically, $N$ is an extremely large number given it is the number of observations in the longest degradation signal, while $K$ is usually a small number since high-dimensional signals are often embedded in a relatively low-dimensional subspace. To address this challenge, we may decompose matrix ${\boldsymbol{C}}$ as follows based on Equation \eqref{eq:residual}:

\begin{equation}\label{eq:cdecomp}
{\boldsymbol{C}}=\left[\boldsymbol{U}\;\;\;\;\tilde{\boldsymbol{x}}\right]=\left[\boldsymbol{U}_{old}\;\;\;\; \frac{\boldsymbol{r}}{\|\boldsymbol{r}\|}\right]\begin{bmatrix}
\boldsymbol{I}       & \hat{\boldsymbol{w}} \\
\boldsymbol{0}        & \|\boldsymbol{r}\|
\end{bmatrix},
\end{equation}

\noindent where $\boldsymbol{I}\in\mathbb{R}^{K\times K}$ is an identity matrix, $\boldsymbol{0}\in\mathbb{R}^{1\times K}$ is a zero vector, $\|\boldsymbol{r}\|\in\mathbb{R}$ is the norm of the residual vector $\boldsymbol{r}$. Equation \eqref{eq:cdecomp} implies that, instead of conducting SVD on ${\boldsymbol{C}}$, we may conduct SVD on $\begin{bmatrix}
\boldsymbol{I}       & \hat{\boldsymbol{w}} \\
\boldsymbol{0}        & \|\boldsymbol{r}\|
\end{bmatrix}$, which is a $(K+1)\times (K+1)$ matrix. Let $\begin{bmatrix}
\boldsymbol{I}       & \hat{\boldsymbol{w}} \\
\boldsymbol{0}        & \|\boldsymbol{r}\|
\end{bmatrix}=\hat{\boldsymbol{U}}\hat{\boldsymbol{\Sigma}}\hat{\boldsymbol{V}}$, where $\hat{\boldsymbol{U}}\in\mathbb{R}^{(K+1)\times (K+1)}$, then the left singular vector matrix of ${\boldsymbol{C}}$ is $\tilde{\boldsymbol{U}}=\left[\boldsymbol{U}\;\; \frac{\boldsymbol{r}}{\|\boldsymbol{r}\|}\right]\hat{\boldsymbol{U}}$. As a result, $\boldsymbol{U}_{new}$ can be yielded by keeping the first $K$ columns of $\tilde{\boldsymbol{U}}$.

\begin{algorithm}
    \textit{User 1} Randomly initializes $\boldsymbol{U}_{new}$, the columns of which span the subspace

    \If {The convergence criterion not met}{
    \For{User $i=1$ to $I$}{
    \For{System $j=1$ to $J_i$}{

     $\boldsymbol{U}_{old}\coloneqq\boldsymbol{U}_{new}$

       Project signal $\boldsymbol{x}_{ij}$ onto the current subspace $\hat{\boldsymbol{w}} \coloneqq \arg\min_{\boldsymbol{w}} \|\boldsymbol{U}_{old}^{\boldsymbol{\Omega}_{ij}}\boldsymbol{w}-\boldsymbol{x}_{ij}^{\boldsymbol{\Omega}_{ij}}\|_{2}^{2}$

     Impute missing entries of the signal $[\tilde{\boldsymbol{x}}_{ij}]_n \coloneqq\begin{cases}
      [{\boldsymbol{x}}_{ij}]_n, & n \in \boldsymbol{\Omega}_{ij}\\
      [\boldsymbol{U}_{old}\hat{\boldsymbol{w}}]_n, & n \in \boldsymbol{\Omega}_{ij}^C\end{cases}$
      
       Compute the residual vector ${\boldsymbol{r}} \coloneqq \tilde{\boldsymbol{x}}_{ij} - \boldsymbol{U}_{old}\hat{\boldsymbol{w}}$
      
      Conduct SVD on $\begin{bmatrix}
        \boldsymbol{I}       & \hat{\boldsymbol{w}} \\
        \boldsymbol{0}        & \|\boldsymbol{r}\|
        \end{bmatrix}=\hat{\boldsymbol{U}}\hat{\boldsymbol{\Sigma}}\hat{\boldsymbol{V}}$

        $\tilde{\boldsymbol{U}}\coloneqq\left[\boldsymbol{U}\;\; \frac{\boldsymbol{r}}{\|\boldsymbol{r}\|}\right]\hat{\boldsymbol{U}}$

        $\boldsymbol{U}_{new}\coloneqq\tilde{\boldsymbol{U}}{(1:N,1:K)}$
        }
        
        Share the new matrix $\boldsymbol{U}_{new}$ with the next user}
        }
        
\caption{Federated Dominant Subspace Identification}
\label{alg:21}
\end{algorithm}


We summarize the proposed Federated Dominant Subspace Identification in Algorithm 1. Similar to some of the matrix completion and incremental singular value decomposition algorithms \cite{brand2002incremental,balzano2010online,dai2010set,keshavan2010matrix}, one of the possible metrics to determine if Algorithm 1 has converged or not is the \textit{residual norm}, which is defined as follows: 

\begin{equation}\label{eq:residualnorm}
e = \sum_{i=1}^{I}\sum_{j=1}^{J_i}\frac{\|\boldsymbol{r}_{ij}\|}{\|\tilde{\boldsymbol{x}}_{ij}\|} = \sum_{i=1}^{I}\sum_{j=1}^{J_i}\frac{\|\boldsymbol{r}_{ij}^{\boldsymbol{\Omega}_{ij}}\|}{\|\boldsymbol{x}_{ij}^{\boldsymbol{\Omega}_{ij}}\|}
\end{equation}

\noindent where $\|\cdot\|$ is the $\ell_2$ norm, and $\boldsymbol{r}_{ij}$ is the residual of projecting vector $\boldsymbol{x}_{ij}$ onto the current subspace (see Row 9 in Algorithm 1). To determine if Algorithm 1 has converged or not, the first user will compute the summation of its residual norm (i.e., $\sum_{j=1}^{J_1}\frac{\|\boldsymbol{r}_{1j}\|}{\|\tilde{\boldsymbol{x}}_{1j}\|}$) and pass it to the second user; the second user will add its residual norm to the norm from the previous user and pass it to next user, so on so forth; the last user will calculate the total residual norm $e$ from all the users. If $e$ is smaller than a predefined small number, say $10^{-6}$, then Algorithm 1 is considered to be converged. Another possible way to terminate Algorithm 1 is the total number of iterations. Specifically, if the total number of iterations exceeds a predefined threshold, say $100$, then Algorithm 1 stops \cite{balzano2013grouse}. 

\subsubsection{Federated Algorithm for MFPC-scores Computation}\label{sec:sub:sub:mfpc}

Algorithm 1 provides a (general) set of basis vectors (i.e., $\boldsymbol{U}_{new}$) for the $K$-dimensional dominant subspace of signal matrix $\boldsymbol{{X}}$. As discussed in Section \ref{sec: sub: para_center}, MFPC-scores are computed using $\boldsymbol{Z}=\boldsymbol{U}_{dominant}^\top\boldsymbol{\tilde{X}}$, which requires the centered signal matrix (i.e., $\boldsymbol{\tilde{X}}$) and its first $K$ left singular vectors (i.e., the columns of $\boldsymbol{U}_{dominant}$). However, both $\boldsymbol{\tilde{X}}$ and $\boldsymbol{U}_{dominant}$ are unknown and cannot be easily computed under a federated learning framework. To address this challenge, we will first show that MFPC-scores can be computed using basis vectors $\boldsymbol{U}_{new}$ and the uncentered signal matrix $\boldsymbol{{X}}$. Next, we will propose a federated algorithm that uses $\boldsymbol{U}_{new}$ and $\boldsymbol{{X}}$ to compute MFPC-scores. This allows multiple users to jointly compute their MFPC-scores while keeping each user's data local and confidential.

To show that MFPC-scores can be computed using basis vectors $\boldsymbol{U}_{new}$ and the uncentered signal matrix $\boldsymbol{{X}}$, we first introduce Proposition 1. 

\begin{prop}
Given an uncentered matrix $\boldsymbol{X}\in\mathbb{R}^{N\times J}$ and $\boldsymbol{U}_{new}\in\mathbb{R}^{N\times K}$ whose orthonormal columns span the $K$-dimensional dominant subspace of $\boldsymbol{X}$, the MFPC-scores discussed in Section \ref{sec: sub: para_center} can be computed from $\boldsymbol{X}$ and $\boldsymbol{U}_{new}$ as follows: 
(i) compute $\boldsymbol{W}=\boldsymbol{U}_{new}^\top\boldsymbol{X}\in\mathbb{R}^{K\times J}$, (ii) center $\boldsymbol{W}$ by computing $\boldsymbol{\tilde{W}}=\boldsymbol{W}-\boldsymbol{\bar{W}}\in\mathbb{R}^{K\times J}$, where $\boldsymbol{\bar{W}}\in\mathbb{R}^{K\times J}$ has $J$ identical columns, each of which is the column mean of $\boldsymbol{{W}}$, (iii) conduct SVD on $\boldsymbol{\tilde{W}}$--that is--$\boldsymbol{\tilde{W}}=\boldsymbol{P}\boldsymbol{D}\boldsymbol{Q}^\top$, where $\boldsymbol{P}\in\mathbb{R}^{K\times K}$ is the left singular vector matrix, and (iv) compute MFPC-scores $\boldsymbol{Z}=\boldsymbol{P}^\top\boldsymbol{\tilde{W}}\in\mathbb{R}^{K\times J}$.

\end{prop}

The proof of Proposition 1 can be found in the Appendix. Geometrically, Proposition 1 can be explained as follows. The columns of $\boldsymbol{U}_{new}\in\mathbb{R}^{N\times K}$ constructs a \textit{new coordinate system} in the $K$-dimensional dominant subspace of $\boldsymbol{X}$, which is different from the \textit{default coordinate system} in which $\boldsymbol{X}$ is defined. Steps (i)-(iv) compute MFPC-scores in the \textit{new coordinate system}. Specifically, step (i), $\boldsymbol{W}=\boldsymbol{U}_{new}^\top\boldsymbol{X}\in\mathbb{R}^{K\times J}$, tries to express the the $J$ data points (a.k.a. samples) in $\boldsymbol{X}\in\mathbb{R}^{N\times J}$ in the \textit{new coordinate system}, and $\boldsymbol{W}\in\mathbb{R}^{K\times J}$ contains the coordinates of the $J$ data points. In step (ii), we center $\boldsymbol{W}$, which yields $\boldsymbol{\tilde{W}}=\boldsymbol{W}-\boldsymbol{\bar{W}}\in\mathbb{R}^{K\times J}$. This is the same as centering the original signal matrix $\boldsymbol{X}$ in the \textit{default coordinate system}. This is because $\boldsymbol{\tilde{W}}=\boldsymbol{U}_{new}^\top\boldsymbol{\tilde{X}}$, where $\boldsymbol{\tilde{X}}=\boldsymbol{X}-\boldsymbol{\bar{X}}$ is the centered signal matrix (see Section \ref{sec: sub: para_center} for definitions). In step (iii), we conduct SVD on $\boldsymbol{\tilde{W}}$, which gives $\boldsymbol{\tilde{W}}=\boldsymbol{P}\boldsymbol{D}\boldsymbol{Q}^\top$, where $\boldsymbol{P}\in\mathbb{R}^{K\times K}$ is the left singular vector matrix. Since $\boldsymbol{\tilde{W}}$ contains the coordinates of the $J$ data points in the \textit{new coordinate system}, the singular vectors in $\boldsymbol{P}$ are defined in the \textit{new coordinate system} as well. Step (iii) is equivalent to conducting SVD on $\boldsymbol{\tilde{X}}$ in the \textit{default coordinate system}, the left $K$ dominant singular vectors of which are the columns of $\boldsymbol{U}_{dominant}\in\mathbb{R}^{N\times K}$. Notice that $\boldsymbol{P}\in\mathbb{R}^{K\times K}$ and $\boldsymbol{U}_{dominant}\in\mathbb{R}^{N\times K}$ are the same set of singular vectors defined in two different coordinate systems, where $\boldsymbol{P}\in\mathbb{R}^{K\times K}$ is defined in the \textit{new coordinate system} constructed by the columns of $\boldsymbol{U}_{new}\in\mathbb{R}^{N\times K}$, while $\boldsymbol{U}_{dominant}\in\mathbb{R}^{N\times K}$ is defined in the \textit{default coordinate system}. The transformation of the two coordinate systems can be easily achieved by using $\boldsymbol{U}_{new}\boldsymbol{P}=\boldsymbol{U}_{dominant}$. Given the singular vectors, step (iv) computes MFPC-scores $\boldsymbol{Z}=\boldsymbol{P}^\top\boldsymbol{\tilde{W}}\in\mathbb{R}^{K\times J}$. The geometric meaning of computing MFPC-scores is that the singular vectors (the columns of $\boldsymbol{P}$) define another coordinate system, and we express the centered $J$ data points $\boldsymbol{\tilde{W}}$ in this coordinate system ($\boldsymbol{P}^\top\boldsymbol{\tilde{W}}$ are the new coordinates) such that the first dimension will have the maximum variance, the second dimension has the secondary maximum variance, so on so forth.

Proposition 1 indicates that MFPC-scores can be computed using basis vectors $\boldsymbol{U}_{new}$ and the uncentered signal matrix $\boldsymbol{{X}}$. Based on Proposition 1, we propose a federated algorithm that allows multiple users to use $\boldsymbol{U}_{new}$ and their uncentered signals to compute MFPC-scores. Following the notations defined earlier, there are $I$ users, and the degradation signal of user $i$'s $j$th system is denoted as $\boldsymbol{x}_{ij}\in\mathbb{R}^{N}$, where $i=1,\ldots, I$ and $j=1,\ldots, J_i$. $I$ is the total number of users, and $J_i$ is the number of systems that user $i$ has. Our proposed federated algorithm requires a server to coordinate the MFPC-scores computation. It can be a standalone server on the cloud or any of the users participating in federated learning. The proposed federated learning algorithm works as follows.

First, each user computes the coordinates of its signals in the \textit{new coordinate systems} constructed by the columns of $\boldsymbol{U}_{new}\in\mathbb{R}^{N\times K}$. The new coordinates of signal $\boldsymbol{x}_{ij}$ is computed using $\boldsymbol{w}_{ij}=\boldsymbol{U}_{new}^\top\boldsymbol{x}_{ij}\in\mathbb{R}^{K}$ if $\boldsymbol{x}_{ij}$ is complete. However, degradation signal observations are often incomplete, so not all the observations of $\boldsymbol{x}_{ij}$ are available. To address this challenge, we follow the suggestion of article \cite{balzano2010high} and compute $\boldsymbol{w}_{ij}$ by solving the following optimization problem $\boldsymbol{w}_{ij}\coloneqq\min_{\boldsymbol{w}_{ij}}{\|\boldsymbol{x}_{ij}^{\boldsymbol{\Omega}_{ij}}-\boldsymbol{U}_{new}^{\boldsymbol{\Omega}_{ij}}\boldsymbol{w}_{ij}\|_2^2}$, which yields the solution $\boldsymbol{w}_{ij}=({\boldsymbol{U}_{new}^{\boldsymbol{\Omega}_{ij}}}^\top\boldsymbol{U}_{new}^{\boldsymbol{\Omega}_{ij}})^{-1}\boldsymbol{U}_{new}^{\boldsymbol{\Omega}_{ij}}\boldsymbol{x}_{ij}^{\boldsymbol{\Omega}_{ij}}$. Here $\boldsymbol{\Omega}_{ij}$ is the set of indices of available observations in $\boldsymbol{x}_{ij}$. This is a reasonable estimator because the elements of vector $\boldsymbol{w}_{ij}$ are the weights when expressing $\boldsymbol{x}_{ij}$ as a linear weighted combination of the columns of $\boldsymbol{U}_{new}$. After computing the weights from all $J_i$ signals (i.e., $\boldsymbol{w}_{i1},\boldsymbol{w}_{i2},\ldots,\boldsymbol{w}_{iJ_i}$), user $i$ will send $\boldsymbol{w}_{(i)}=[\boldsymbol{w}_{i1},\boldsymbol{w}_{i2},\ldots,\boldsymbol{w}_{iJ_i}]\in\mathbb{R}^{K\times J_i}$ to the server. 

Second, the server receives $\boldsymbol{w}_i$ from all $I$ users and integrates them into matrix $\boldsymbol{W}=[\boldsymbol{w}_{(1)}, \boldsymbol{w}_{(2)},\ldots,\boldsymbol{w}_{(I)}]\in\mathbb{R}^{K\times J}$, where $J=\sum_{i=1}^IJ_i$. Then, MFPC-scores can be computed by following steps (ii)-(iv) in Proposition 1. Specifically, the server first centers $\boldsymbol{W}$ by computing $\boldsymbol{\tilde{W}}=\boldsymbol{W}-\boldsymbol{\bar{W}}\in\mathbb{R}^{K\times J}$, where $\boldsymbol{\bar{W}}\in\mathbb{R}^{K\times J}$ has $J$ identical columns, each of which is $\bar{\boldsymbol{w}}=\sum_{i=1}^{I}\sum_{j=1}^{J_i}\boldsymbol{w}_{ij}\in\mathbb{R}^{K}$. Next, it conducts SVD on $\boldsymbol{\tilde{W}}$--that is--$\boldsymbol{\tilde{W}}=\boldsymbol{P}\boldsymbol{D}\boldsymbol{Q}^\top$, where $\boldsymbol{P}\in\mathbb{R}^{K\times K}$ is the left singular vector matrix. Finally, MFPC-scores are computed by using $\boldsymbol{Z}=\boldsymbol{P}^\top\boldsymbol{\tilde{W}}\in\mathbb{R}^{K\times J}$. Matrix $\boldsymbol{Z}$ consists of the MFPC-scores of all $I$ users, where the first $J_1$ columns are the MFPC-scores of user $1$ (denoted as $\boldsymbol{z}_1\in\mathbb{R}^{K\times J_1}$), the following $J_2$ columns are the scores of user $2$ (denoted as $\boldsymbol{z}_2\in\mathbb{R}^{K\times J_2}$), so on so forth. In other words, $\boldsymbol{Z}=[\boldsymbol{z}_1\in\mathbb{R}^{K\times J_1},\boldsymbol{z}_2\in\mathbb{R}^{K\times J_2},\ldots,\boldsymbol{z}_I\in\mathbb{R}^{K\times J_I}]$. We summarize the proposed federated algorithm for MFPC-scores computation in Algorithm 2.


\begin{algorithm}
{\bf{\textit{User side}}}

\For{User $i=1$ to $I$}{ 
\For{$j=1$ to $J_i$}{
Compute $\boldsymbol{w}_{ij}=({\boldsymbol{U}_{new}^{\boldsymbol{\Omega}_{ij}}}^\top\boldsymbol{U}_{new}^{\boldsymbol{\Omega}_{ij}})^{-1}\boldsymbol{U}_{new}^{\boldsymbol{\Omega}_{ij}}\boldsymbol{x}_{ij}^{\boldsymbol{\Omega}_{ij}}\in\mathbb{R}^{K}$
}
Merge $\boldsymbol{w}_{(i)}=[\boldsymbol{w}_{i1},\boldsymbol{w}_{i2},\ldots,\boldsymbol{w}_{iJ_i}]\in\mathbb{R}^{K\times J_i}$ and send it to server
}

{\bf{\textit{Server side}}}

Integrate $\boldsymbol{w}_{(i)}$ from all users, which yields $\boldsymbol{W}=[\boldsymbol{w}_{(1)}, \boldsymbol{w}_{(2)},\ldots,\boldsymbol{w}_{(I)}]\in\mathbb{R}^{K\times J}$, where $J=\sum_{i=1}^IJ_i$.

Center $\boldsymbol{W}$ by computing $\boldsymbol{\tilde{W}}=\boldsymbol{W}-\boldsymbol{\bar{W}}\in\mathbb{R}^{K\times J}$, where $\boldsymbol{\bar{W}}\in\mathbb{R}^{K\times J}$ has $J$ identical columns, each of which is $\bar{\boldsymbol{w}}=\sum_{i=1}^{I}\sum_{j=1}^{J_i}\boldsymbol{w}_{ij}\in\mathbb{R}^{K}$.

Conduct SVD on $\boldsymbol{\tilde{W}}$, i.e., $\boldsymbol{\tilde{W}}=\boldsymbol{P}\boldsymbol{D}\boldsymbol{Q}^\top$, where $\boldsymbol{P}\in\mathbb{R}^{K\times K}$ is the left singular vector matrix. 

Compute MFPC-scores $\boldsymbol{Z}=\boldsymbol{P}^\top\boldsymbol{\tilde{W}}\in\mathbb{R}^{K\times J}$.

Send $\boldsymbol{P}$, $\bar{\boldsymbol{w}}$, and each user's MFPC-scores back.

\caption{Federated Algorithm for MFPC-scores Computation}
\label{alg:2}
\end{algorithm}

Notice that user $i$ sends $\boldsymbol{w}_{(i)}$ instead of its signals $\boldsymbol{x}_{ij}^{\boldsymbol{\Omega}_{ij}}$ to the server. Since $\boldsymbol{w}_{ij}=({\boldsymbol{U}_{new}^{\boldsymbol{\Omega}_{ij}}}^\top\boldsymbol{U}_{new}^{\boldsymbol{\Omega}_{ij}})^{-1}\boldsymbol{U}_{new}^{\boldsymbol{\Omega}_{ij}}\boldsymbol{x}_{ij}^{\boldsymbol{\Omega}_{ij}}$ and $\boldsymbol{\Omega}_{ij}$ is confidential (user $i$'s $\boldsymbol{\Omega}_{ij}$ is not shared with other users), signal $\boldsymbol{x}_{ij}^{\boldsymbol{\Omega}_{ij}}$ cannot be derived from $\boldsymbol{w}_{ij}$. This protects the data privacy of each user's data.

\subsection{Prognostic Model Training and Time-To-Failure Prediction}\label{sec:sub:prediction}

After executing Algorithms \ref{alg:21} and \ref{alg:2}, each user $i$, where $i=1,\ldots,I$, receives its MFPC-scores $\boldsymbol{z}_i\in\mathbb{R}^{K\times J_i}$, along with the mean weight vector $\bar{\boldsymbol{w}}\in\mathbb{R}^{K}$ and the singular vectors $\boldsymbol{P}\in\mathbb{R}^{K\times K}$ from the server. Additionally, recall that each user $i$ also possesses the basis of the dominant subspace $\boldsymbol{U}_{new}\in\mathbb{R}^{N\times K}$ and the TTFs (time-to-failure) of its local training data, denoted as $\boldsymbol{y}_i\in\mathbb{R}^{J_i}$.

Utilizing their MFPC-scores $\{\boldsymbol{z}_i\}_{i=1}^{I}$ and TTFs $\{\boldsymbol{y}_i\in\mathbb{R}^{J_i}\}_{i=1}^{I}$, all the $I$ users can jointly train a prognostic model based on (log)-location-scale (LLS) regression. In this article, we will employ the federated LLS regression model developed in \cite{su2022}, which allows the $I$ users to collaboratively train the prognostic model while keeping each user's MFPC-scores and TTFs local and confidential. The parameter estimation of the federated LLS regression utilizes maximum likelihood estimation, which is solved using gradient descent. Initially, the server randomly initializes the regression coefficients. Subsequently, each user downloads these initial coefficients and computes its local gradients using its local data (i.e., MFPC-scores and TTFs). These local gradients are then transmitted to the server, which aggregates them from all users and employs the aggregated gradients to update the regression coefficients. This updated set of coefficients is subsequently redistributed to each user for another round of updating, repeating this process until convergence is achieved. The federated LLS regression provides each user with the estimated regression parameters, which we denoted as $\hat{\beta}_0\in\mathbb{R}$, $\hat{\boldsymbol{\beta}}\in\mathbb{R}^K$, and $\hat\sigma\in\mathbb{R}$ (please refer to Equation \eqref{eq5}).

With the trained federated LLS, we can predict the TTF of an in-field partially degraded system using its real-time (incomplete or complete) degradation signals. Assume user $i$ has a partially degraded system whose real-time degradation signal is denoted as $\boldsymbol{x}_{i, r}^{\boldsymbol{\Omega}_{i, r}}$. To predict the system's TTF, we first compute $\boldsymbol{w}_{i, r}=({\boldsymbol{U}_{new}^{\boldsymbol{\Omega}_{i, r}}}^\top\boldsymbol{U}_{new}^{\boldsymbol{\Omega}_{i, r}})^{-1}\boldsymbol{U}_{new}^{\boldsymbol{\Omega}_{i, r}}\boldsymbol{x}_{i, r}^{\boldsymbol{\Omega}_{i, r}}\in\mathbb{R}^{K}$. Next, MFPC-scores can be computed $\boldsymbol{z}_{i,r}=\boldsymbol{P}^\top(\boldsymbol{w}_{i, r}-\bar{\boldsymbol{w}})$. Finally, $\boldsymbol{z}_{i,r}$ is fed into the trained prognostic model, and TTF can be predicted: $\hat y_{ir}\sim LLS(\hat{\beta}_0+\hat{\boldsymbol{\beta}}^\top\boldsymbol{z}_{i,r},\hat\sigma)$.

One crucial aspect that requires discussion is the determination of $K$ during the execution of Algorithms \ref{alg:21} and \ref{alg:2} when training the prognostic model. $K$ is the dimension of the subspace, which is also the number of principal components to be kept in MFPCA. As mentioned earlier, the value of $K$ can be selected using Fraction-of-Variance Explained (FVE) \cite{fang2015adaptive} or cross-validation (CV) \cite{fang2017multistream}. If FVE is chosen, we usually set the dimension of the subspace as a large number in Algorithm \ref{alg:21}. Recall that the signal matrix  $\boldsymbol{{X}}\in\mathbb{R}^{N\times J}$, then the dimension of the subspace can be set as $\min\{N,J\}$ by the first user. After executing Algorithm \ref{alg:2}, we can use the singular values (the diagonal elements of $\boldsymbol{D}$ in the $11$th row of Algorithm \ref{alg:2}) to compute FVE and determine the minimum number of principal components that can explain enough information of the signal matrix $\boldsymbol{{X}}$ as follows: $K=\min_j\sum_{i=1}^jd_i^2/\sum_{i}d_i^2\geq T_{FVE}$. Here, $\sum_{i=1}^j d_i^2$ denotes the sum of the squares of the first $j$ diagonal elements of matrix $\boldsymbol{D}$, while $\sum_{i} d_i^2$ represents the sum of the squares of all diagonal elements of $\boldsymbol{D}$. $T_{FVE}$ is a threshold between $0$ and $1$, which is usually set as a number close to $1$, say $0.9$.

If $M$-fold cross-validation is utilized to determine $K$, each user initially randomly partitions its local data into $M$ folds. Subsequently, a range of different $K$ values is explored. For each $K$ value, every user employs $M-1$ folds of its data to execute Algorithms \ref{alg:21} and \ref{alg:2} and train the federated LLS regression model. Then, the remaining fold of data is employed as test data to assess the prediction accuracy of the trained federated LLS regression model. This process iterates $M$ times, ensuring that each fold serves as the test data once. Each user transmits its prediction accuracy from these $M$ iterations to the server, which computes an average prediction accuracy. Following the exploration of all candidate $K$ values, each value is associated with a prediction accuracy, and the $K$ value yielding the lowest prediction accuracy is selected as the optimal $K$. Please note that some users may possess fewer than $M$ data samples. Despite this, these users can still engage in the cross-validation process by randomly excluding certain rounds of training and testing. Alternatively, if the number of such users is minimal, another approach is to omit them when performing cross-validation.

\section{Simulation Study I}
\label{sec:sim}

In this section, we conduct a series of simulation studies to validate the capabilities of our proposed federated prognostic model. 

\subsection{Data Generation and Benchmarking Models}
We generate degradation signals and the TTFs of $120$ systems and assign $54$ systems to user 1, $27$ to user 2, and $9$ to user 3. The remaining $30$ systems are considered test data. Similar to \cite{fang2021multi}, we will first generate the underlying degradation trajectories, based on which TTFs are computed. Next, we will generate degradation signals, which are noisy discrete observations of underlying degradation trajectories. Specifically, the underlying degradation trajectory of system $i$ is simulated using the equation below:

\begin{equation}\label{eq:underlying}
    s_{i}(t)=-c_{i}/\ln{(t)},
\end{equation} 
where  $c_{i}\sim N(1,0.25)$, $0\leq t<1$, $i=1,\ldots,100$. The TTFs are computed as the first time that $s_{i}(t)$ reaches or crosses a predefined threshold $D$: $s_{i}(t)=-c_{i}/\ln{(\tilde{y_i})}=D$. This yields $\ln{(\tilde{y_i})}=-c_i/D$, where $\tilde{y_i}$ is the TTF of system $i$. To mimic data acquisition errors, we add noise to the true TTFs. As a result, the observed TTFs are computed from $\ln{(\tilde{y_i})}=-c_i/D+\epsilon_{i}$, where $\epsilon_{i} \sim N(0,0.025^2)$. It can be easily shown that the relationship between TTFs and degradation signals is lognormal, and thus the lognormal regression will be employed in this simulation study to verify the performance of the proposed federated prognostic model. With the underlying degradation trajectory, the noisy discrete observations (i.e., the observed degradation signal from a condition monitoring sensor) of system $i$, $i=1,\ldots, 100$, are generated as follows:

\begin{equation}\label{eq:underlying}
    {x}(\tau_i)=-c_{i}/\ln{(\tau_i)}+ {\varepsilon}(\tau_i),
\end{equation} 

\noindent where ${\varepsilon}(\tau_i)\sim N(0,0.2)$ is the random observation noise, $\tau_i=0,1\times 15e^{-4},2\times 15e^{-4},3\times 15e^{-4},\ldots,\lfloor\tilde{y_i}/15e^{-4}\rfloor\times 15e^{-4}$, and $\lfloor\cdot\rfloor$ is the floor function.

We compare the performance of the proposed federated prognostic model (designated as ``Federated Model") with two benchmarking methods. The first baseline is referred to as the ``Non-Federated Model," which works by first aggregating the degradation data from all the users. Then, the prognostic model proposed in \cite{fang2021multi} is applied to the aggregated data for failure time prediction. To the best of our knowledge, the model in \cite{fang2021multi} is the prognostic model that is closest to the model proposed in this article. It has three steps: matrix completion for missing data imputation, functional principal component analysis for data fusion, and LLS regression for failure time prediction. However, the model in \cite{fang2021multi} is not federated and thus cannot protect users' data privacy. The second baseline, which we refer to as the ``Individual Model," means each user trains its own prognostic model using the data it has. In other words, the data from different users are not used to jointly train the prognostic model. Specifically, each user uses its incomplete local date to conduct MFPCA and extract low-dimensional features. This is achieved by first running Algorithm 1 proposed in this article to identify the dominant subspace of its local data. Next, Algorithm 2 proposed in this article is employed to compute the MFPC-scores. It's important to note that Algorithm 1 can still operate even with only one user by setting the total number of users to one ($I=1$) within Algorithm 1. Despite the single-user scenario, Algorithm 1 remains iterative, where only the data from that user is involved in each round of updating. Similarly, the same principle applies to Algorithm 2, where $I=1$ is set to accommodate the single-user situation. Running Algorithms 1 and 2 allows each user to use its local data to compute its local MFPC-scores. Given that each user possesses its own local TTFs, they can individually conduct regression analyses of their local TTFs against their respective local MFPC-scores using LLS regression. The regression coefficients can then be estimated through maximum likelihood estimation (MLE).

We consider three levels of data incompleteness, $30\%$, $50\%$, and $70\%$, where, for example, $30\%$ means that $30\%$ observations of the generated degradation signals are randomly removed. Under each incompleteness level, we employ 15 random permutations of the data to evaluate the performance of the proposed method as well as the benchmarks. The rank of the dominant subspace in the proposed method and the number of MFPC-scores in all the models are chosen using 5-fold cross-validation. The predictor errors are computed using the following formula:

\begin{equation}
    \text{Prediction Error} = \frac{|\text{Predicted TTF - Real TTF}|}{\text{Real TTF}}\times 100\%.
\end{equation}

\subsection{Results and Analysis}\label{sim:1}

\begin{figure}[htp]
\begin{center}
    \begin{subfigure}[b]{0.4\linewidth}
         \centering
         \includegraphics[width=\textwidth]{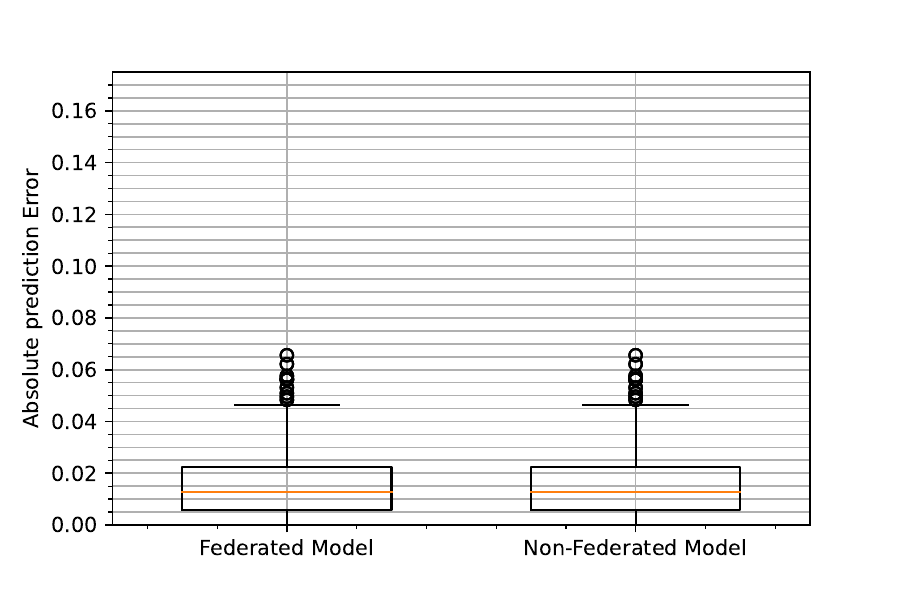}
         \caption{30\% missing values}
         \label{fig:y equals x}
     \end{subfigure}
     \hfill
     \begin{subfigure}[b]{0.4\linewidth}
         \centering
         \includegraphics[width=\textwidth]{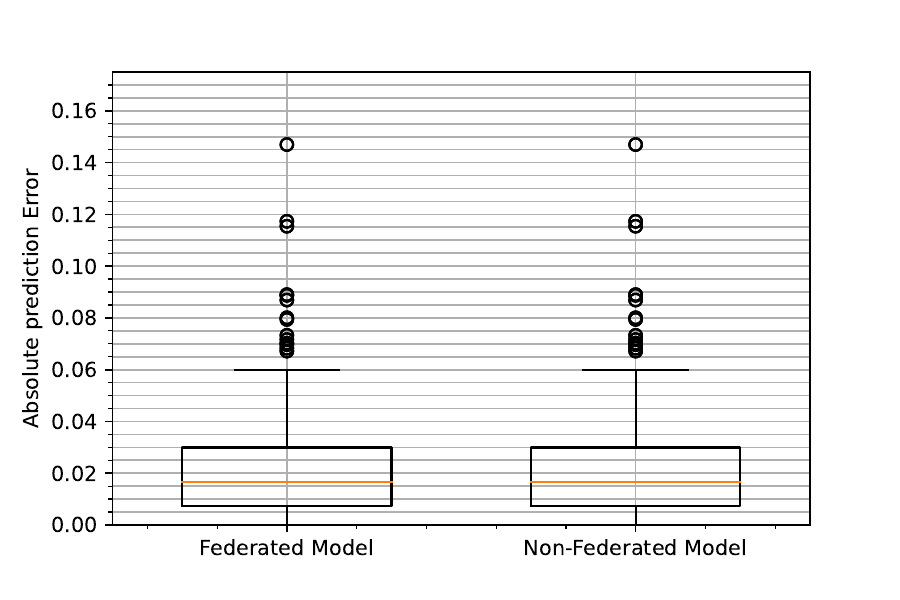}
         \caption{50\% missing values}
         \label{fig:three sin x}
     \end{subfigure}
     \hfill
     \begin{subfigure}[b]{0.4\linewidth}
         \centering
         \includegraphics[width=\textwidth]{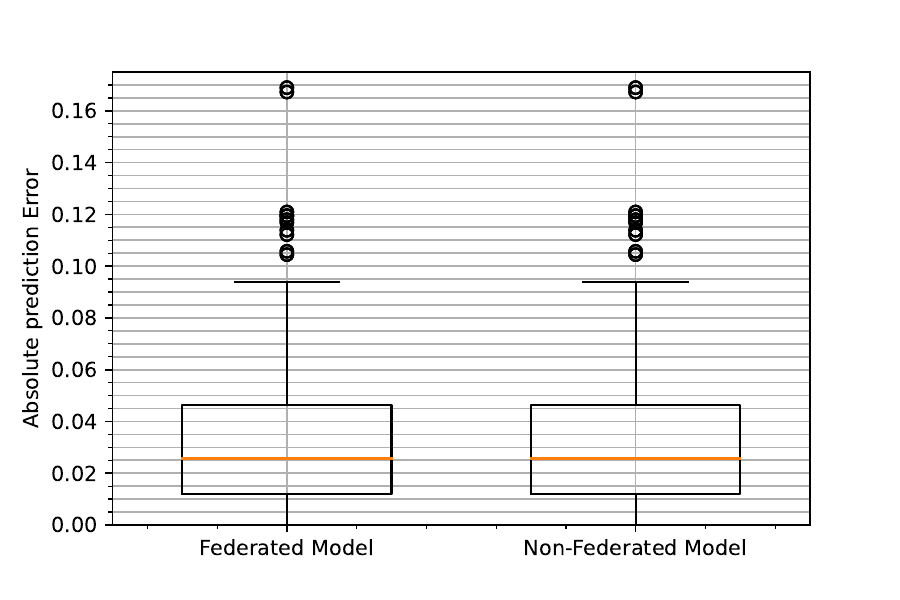}
         \caption{70\% missing values}
         \label{fig:five over x}
     \end{subfigure}
\end{center}
\vspace*{-7mm}
\caption{The prediction errors of federated and non-federated models. \label{fig:sim:federatedvsnonfederated}}
\end{figure}

Figure \ref{fig:sim:federatedvsnonfederated} summarizes the prediction errors of the proposed federated model and the first baseline, which is a non-federated model that works on the aggregated data. It can be seen from Figure \ref{fig:sim:federatedvsnonfederated} that the proposed federated model achieves the same prediction accuracy as the non-federated benchmark at all data missing levels. For example, when $30\%$ observations of degradation signals are missing, the median (and the Interquartile Range, i.e., IQR) of the federated model and non-federated model are both $0.013 (0.016)$; when the data missing level is $50\%$, the median (and IQR) of both models are $0.017 (0.023)$; furthermore, the median (and IQR) are $0.026(0.034)$ for both models when the data missing level is $70\%$. This implies that the proposed federated method does not compromise the prediction performance of prognostic models. Another observation from Figure \ref{fig:case:federatedvsnonfederated} is that the performance of the proposed model and the first benchmark degrades as the data missing level increases. This is reasonable since a higher level of data incompleteness means more data observations are missing, which results in a more inaccurate feature extraction, parameter estimation, and thus TTF prediction. 

\begin{figure}[htp]
\begin{center}
    \begin{subfigure}[b]{0.4\linewidth}
         \centering
         \includegraphics[width=\textwidth]{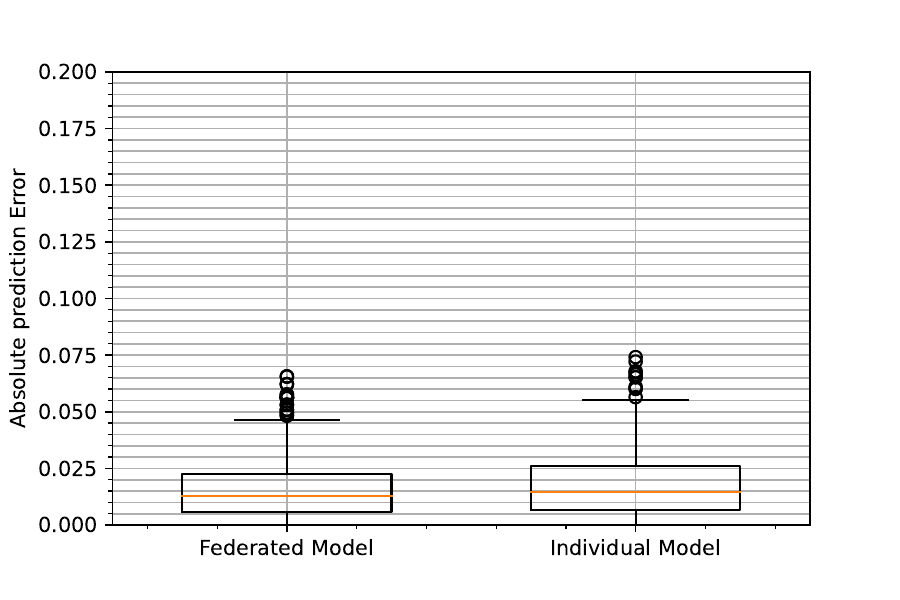}
         \caption{30\% missing values}
         \label{fig:y equals x}
     \end{subfigure}
     \hfill
     \begin{subfigure}[b]{0.4\linewidth}
         \centering
         \includegraphics[width=\textwidth]{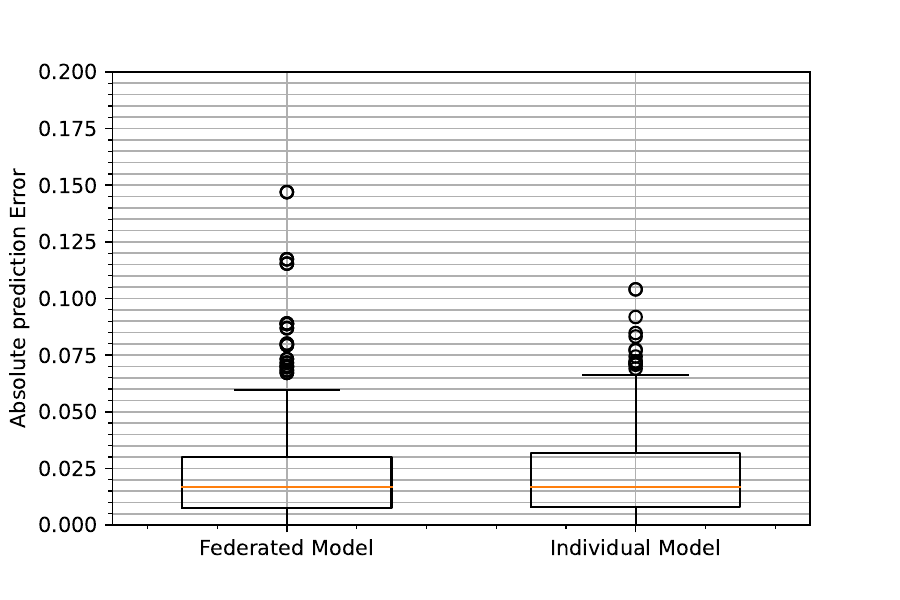}
         \caption{50\% missing values}
         \label{fig:three sin x}
     \end{subfigure}
     \hfill
     \begin{subfigure}[b]{0.4\linewidth}
         \centering
         \includegraphics[width=\textwidth]{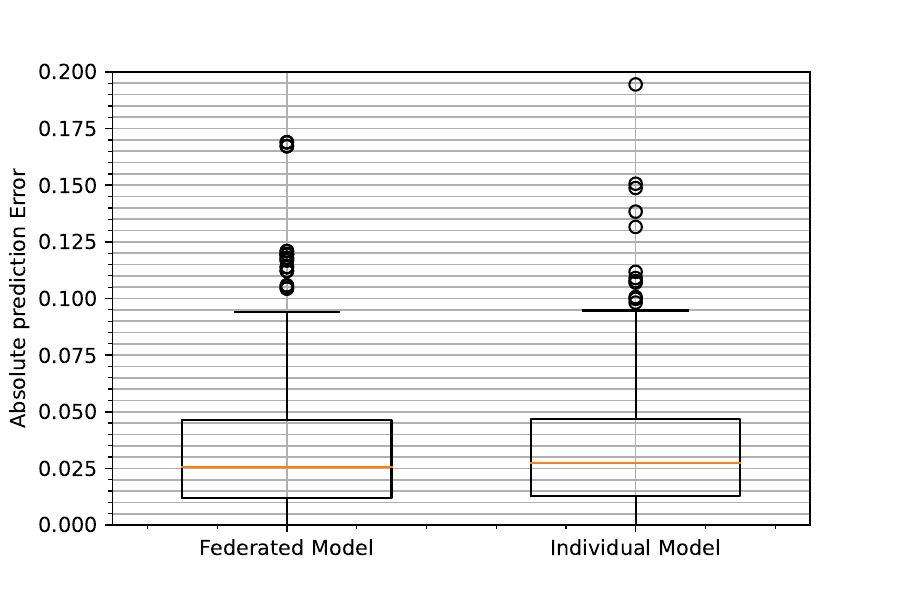}
         \caption{70\% missing values}
         \label{fig:five over x}
     \end{subfigure}
\end{center}
\vspace*{-7mm}
\caption{The prediction errors of federated model and individual model (User 1)\label{fig:sim:user1}}
\end{figure}

\begin{figure}[htp]
\begin{center}
    \begin{subfigure}[b]{0.4\linewidth}
         \centering
         \includegraphics[width=\textwidth]{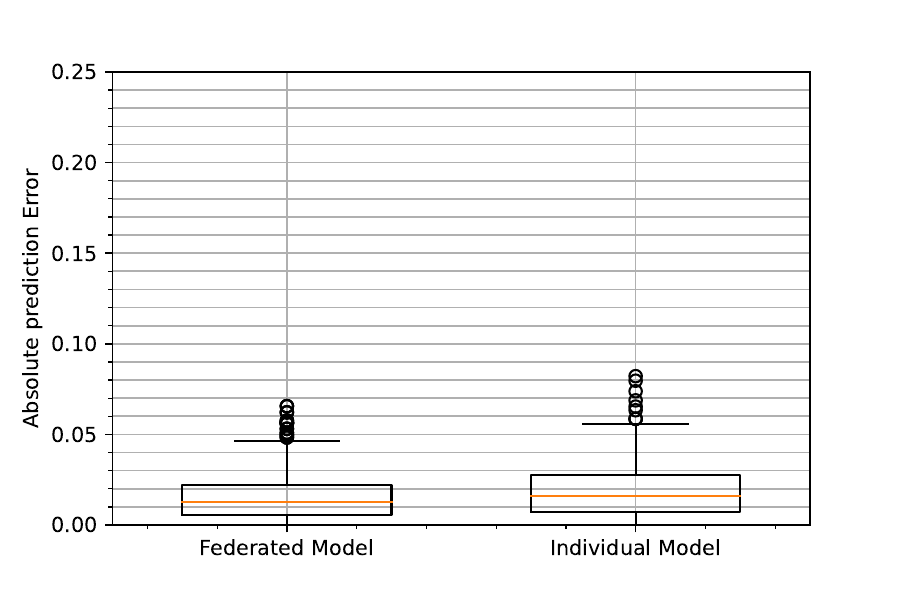}
         \caption{30\% missing values}
         \label{fig:y equals x}
     \end{subfigure}
     \hfill
     \begin{subfigure}[b]{0.4\linewidth}
         \centering
         \includegraphics[width=\textwidth]{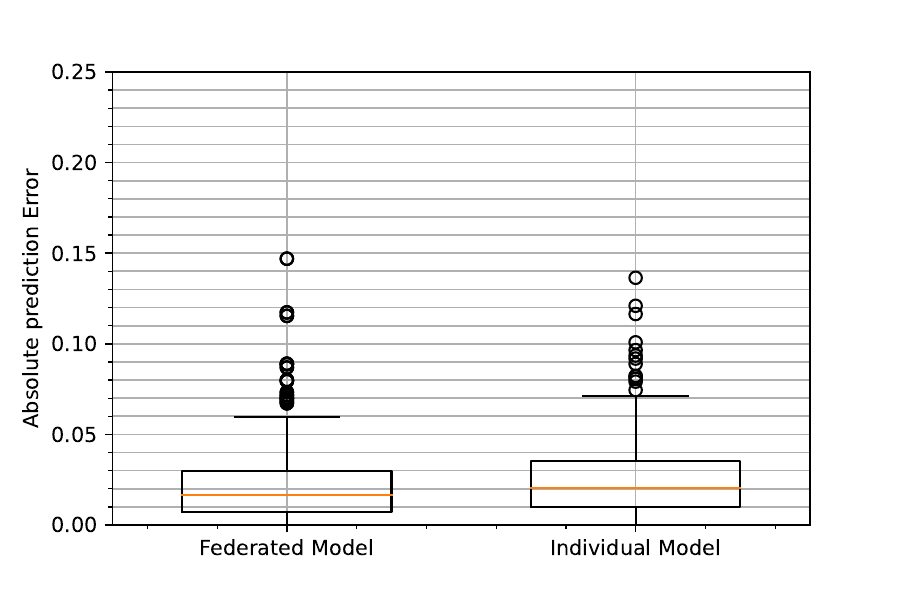}
         \caption{50\% missing values}
         \label{fig:three sin x}
     \end{subfigure}
     \hfill
     \begin{subfigure}[b]{0.4\linewidth}
         \centering
         \includegraphics[width=\textwidth]{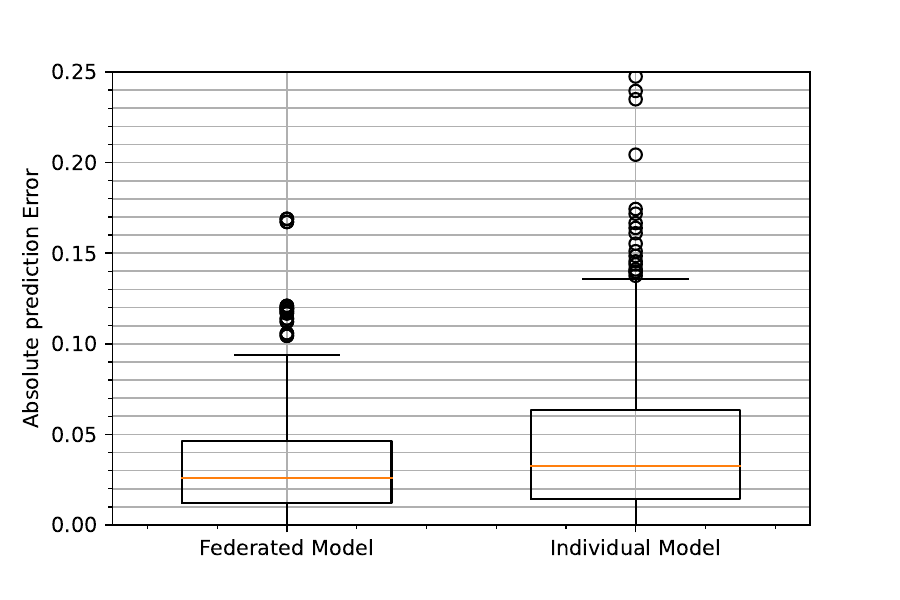}
         \caption{70\% missing values}
         \label{fig:five over x}
     \end{subfigure}
\end{center}
\vspace*{-7mm}
\caption{The prediction errors of federated model and individual model (User 2)\label{fig:sim:user2}}
\end{figure}

\begin{figure}[htp]
\begin{center}
    \begin{subfigure}[b]{0.4\linewidth}
         \centering
         \includegraphics[width=\textwidth]{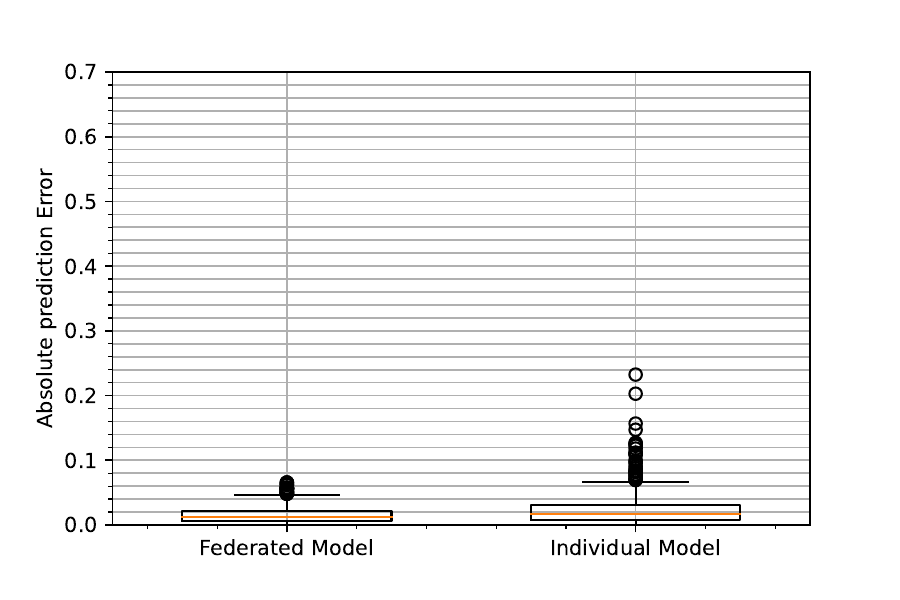}
         \caption{30\% missing values}
         \label{fig:y equals x}
     \end{subfigure}
     \hfill
     \begin{subfigure}[b]{0.4\linewidth}
         \centering
         \includegraphics[width=\textwidth]{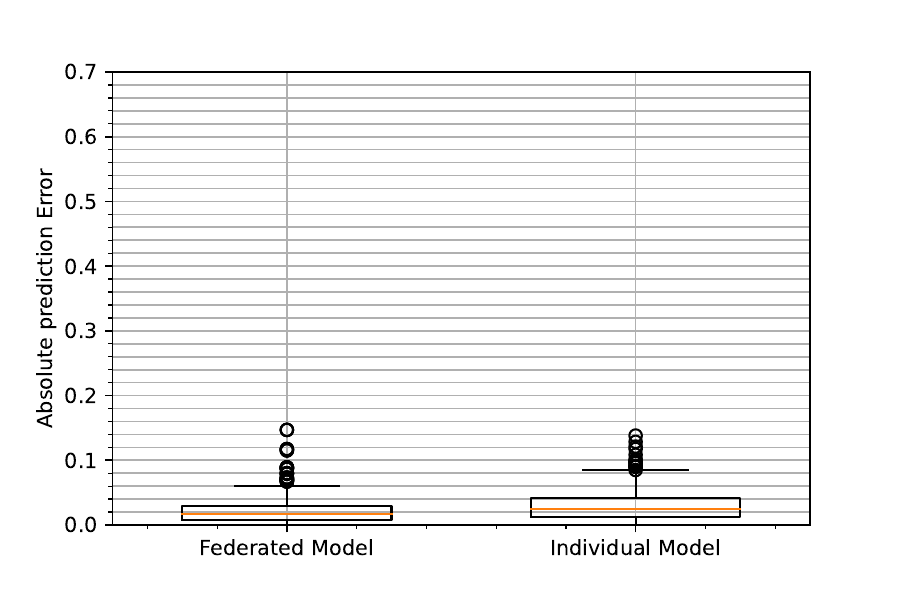}
         \caption{50\% missing values}
         \label{fig:three sin x}
     \end{subfigure}
     \hfill
     \begin{subfigure}[b]{0.4\linewidth}
         \centering
         \includegraphics[width=\textwidth]{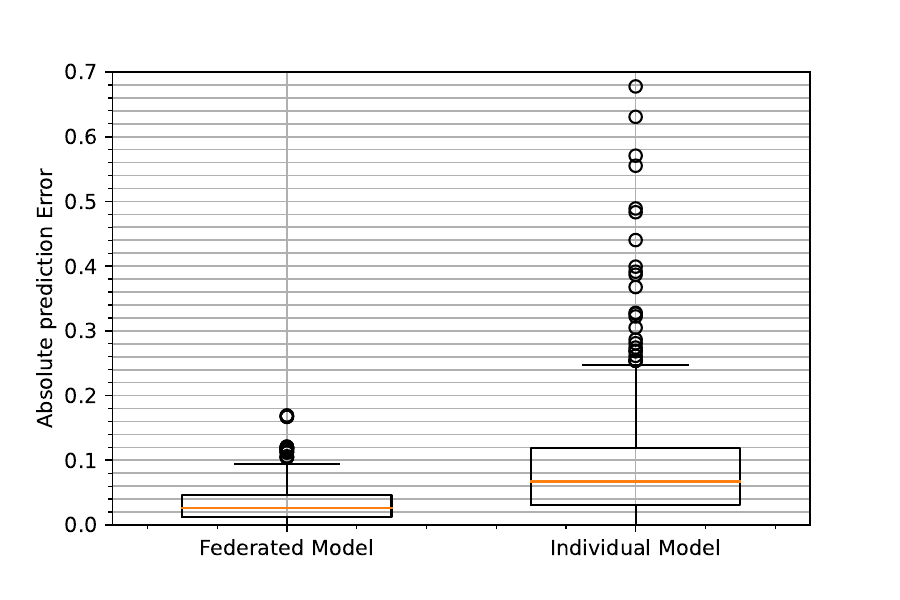}
         \caption{70\% missing values}
         \label{fig:five over x}
     \end{subfigure}
\end{center}
\vspace*{-7mm}
\caption{The prediction errors of federated model and individual model (User 3) \label{fig:sim:user3}}
\end{figure}

Figures \ref{fig:sim:user1}, \ref{fig:sim:user2}, and \ref{fig:sim:user3} report the prediction errors of the proposed federated model and the second benchmark, i.e., the individual model. We can conclude from Figures \ref{fig:sim:user1}, \ref{fig:sim:user2}, and \ref{fig:sim:user3} that the federated model outperforms individual models at all levels of data incompleteness. For example, it can be seen in Figure \ref{fig:sim:user1} that the median (and IQR) of prediction errors of the proposed federated model and the individual model constructed by User 1 are $0.013 (0.016)$ and $0.015 (0.019)$ respectively when data missing level is $30\%$; when $50\%$ data are missing, they are respectively $0.017(0.023)$ and $0.017(0.0.24)$; when data missing level is $70\%$, they are $0.026(0.034)$ and $0.027(0.034)$, respectively. As another example, in Figure \ref{fig:sim:user3}, the median (and IQR) of the proposed federated model and the individual model built by User 3 are $0.013(0.016)$ and $0.017(0.024)$, $0.017(0.023)$ and $0.025(0.029)$, and $0.026(0.034)$ and $0.067(0.089)$ when data missing levels are $30\%$, $50\%$, and $70\%$, respectively. A similar observation can also be found in Figure \ref{fig:sim:user2} as well. This is reasonable since the proposed federated model uses samples from all three users for model training, while the individual models use a smaller number of training samples owned by each individual user. 

Another conclusion can be drawn from Figures \ref{fig:sim:user1}, \ref{fig:sim:user2}, and \ref{fig:sim:user3} is that the performance of individual models constructed by each user is significantly affected by the number of training samples owned by the user. For example, when the data missing level is $30\%$, the median (and IQR) of prediction errors of the individual model constructed by User 1 (Figure \ref{fig:sim:user1}), User 2 (Figure \ref{fig:sim:user2}), and User 3 (Figure \ref{fig:sim:user3}) are respectively $0.015 (0.019)$, $0.016 (0.020)$, and $0.017(0.024)$. This is reasonable since the sample sizes of Users 1, 2, and 3 are $54$, $27$, and $9$, respectively, and when the training sample size is relatively limited, a larger number of training data usually help improve the model performance. This confirms the importance of combining training data from multiple users for a joint prognostic model construction, which also affirms the value of federated failure time prediction. 


\newpage

\section{Simulation Study II}\label{sec:sim2}

In this section, we begin by conducting simulation studies to assess the impact of stragglers on the proposed federated prognostic method. Subsequently, we examine the performance of our method in large-scale applications. Finally, we explore the computational time required by the proposed method.

\subsection{Stragglers Effect}

Federated learning algorithms typically operate iteratively, involving multiple rounds of updates until convergence is achieved. However, not all users may participate in every round of updating, as some users might exhibit slower response times, intermittent disconnections, or choose to exit the federated learning process. This phenomenon is typically referred to as ``stragglers." In this section, we will investigate the effect of stragglers on the performance of the proposed federated industrial prognostic method. Recall that our proposed federated feature extraction method includes two algorithms: federated dominant subspace detection and MFPC-score computation. The first one is an iterative algorithm that requires multiple rounds of updating, while the second algorithm is not iterative and only requires one round of computation. As a result, our focus is on testing the impact of ``stragglers" on the first algorithm. Of course, there is a possibility of encountering stragglers during the execution of the second algorithm. In other words, some users may exit the federated learning process while the second algorithm is in progress. However, its effect on the performance of the proposed industrial prognostic model is obvious as it aligns with reducing the number of training samples for feature extraction and the subsequent federated LLS regression.

\begin{figure}[htp]
\begin{center}
    \begin{subfigure}[b]{0.4\linewidth}
         \centering
         \includegraphics[width=\textwidth]{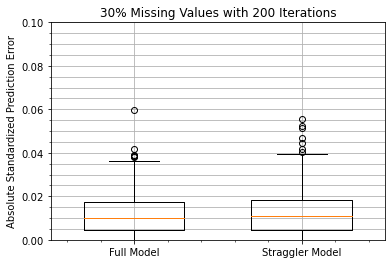}
         \caption{200 iterations}
         \label{fig:y equals x}
     \end{subfigure}
     \hfill
     \begin{subfigure}[b]{0.4\linewidth}
         \centering
         \includegraphics[width=\textwidth]{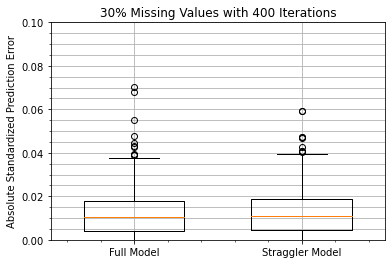}
         \caption{400 iterations}
         \label{fig:three sin x}
     \end{subfigure}
     \hfill
     \begin{subfigure}[b]{0.4\linewidth}
         \centering
         \includegraphics[width=\textwidth]{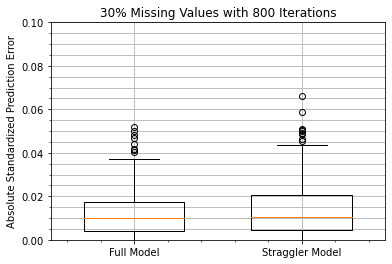}
         \caption{800 iterations}
         \label{fig:five over x}
     \end{subfigure}
\end{center}
\vspace*{-7mm}
\caption{The prediction errors of federated model and straggler model (30\% missing values)\label{fig:sim:stragg1}}
\end{figure}

\begin{figure}[htp]
\begin{center}
    \begin{subfigure}[b]{0.4\linewidth}
         \centering
         \includegraphics[width=\textwidth]{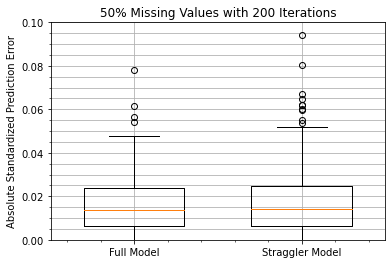}
         \caption{200 iterations}
         \label{fig:y equals x}
     \end{subfigure}
     \hfill
     \begin{subfigure}[b]{0.4\linewidth}
         \centering
         \includegraphics[width=\textwidth]{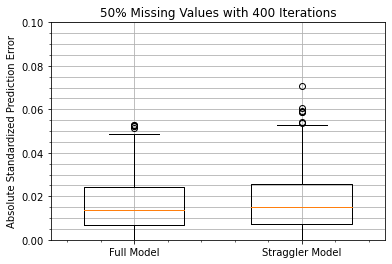}
         \caption{400 iterations}
         \label{fig:three sin x}
     \end{subfigure}
     \hfill
     \begin{subfigure}[b]{0.4\linewidth}
         \centering
         \includegraphics[width=\textwidth]{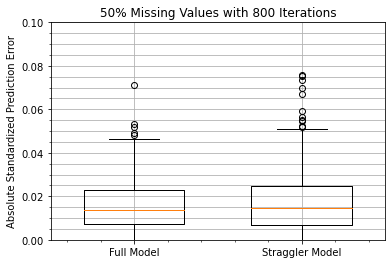}
         \caption{800 iterations}
         \label{fig:five over x}
     \end{subfigure}
\end{center}
\vspace*{-7mm}
\caption{The prediction errors of federated model and straggler model (50\% missing values)\label{fig:sim:stragg2}}
\end{figure}

\begin{figure}[htp]
\begin{center}
    \begin{subfigure}[b]{0.4\linewidth}
         \centering
         \includegraphics[width=\textwidth]{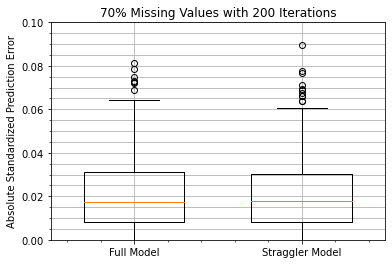}
         \caption{200 iterations}
         \label{fig:y equals x}
     \end{subfigure}
     \hfill
     \begin{subfigure}[b]{0.4\linewidth}
         \centering
         \includegraphics[width=\textwidth]{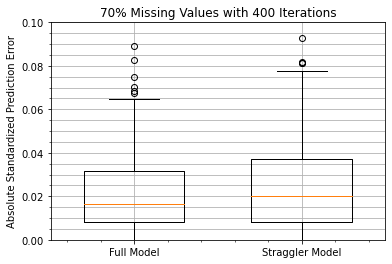}
         \caption{400 iterations}
         \label{fig:three sin x}
     \end{subfigure}
     \hfill
     \begin{subfigure}[b]{0.4\linewidth}
         \centering
         \includegraphics[width=\textwidth]{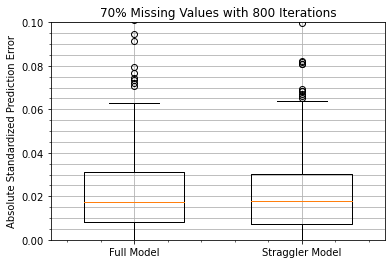}
         \caption{800 iterations}
         \label{fig:five over x}
     \end{subfigure}
\end{center}
\vspace*{-7mm}
\caption{The prediction errors of federated model and straggler model (70\% missing values)\label{fig:sim:stragg3}}
\end{figure}

In this section, the data is simulated using the similar settings discussed in Section \ref{sec:sim}. Here, we generate data for 5 users, where each of them has $33$, $70$, $55$, $56$, and $67$ training samples, respectively. In addition to the training data, we generate another $30$ samples for model testing. To evaluate the impact of stragglers on the proposed method, in each iteration, we randomly select one user as the straggler and exclude it from the model training process. We refer to this model as the ``Stragglers Model" and compare its performance to the proposed federated model without stragglers, which is designated as the ``Full Model." Given Algorithm 1 is iterative, we evaluate the ``Stragglers Model" and ``Full Model" under three different numbers of iterations: 200, 400, and 800. The entire process of data generation and evaluation is repeated $10$ times. Figures \ref{fig:sim:stragg1}, \ref{fig:sim:stragg2}, and \ref{fig:sim:stragg3} illustrate the prediction error of the two models when $30\%$, $50\%$, and $70\%$ observations are missing, respectively.

Figures \ref{fig:sim:stragg1}, \ref{fig:sim:stragg2}, and \ref{fig:sim:stragg3} indicate that the performance of the straggler model is comparable to the full model across all levels of data incompleteness, especially when the number of iteration is large. We believe this is reasonable, and the rationale is outlined as follows. Recall that the federated dominant subspace detection algorithm (i.e., Algorithm 1) is iterative, which is a multiple-round updating algorithm. In the ``Full Model" without stragglers, in each iteration, every user utilizes its data to update the subspace. In contrast, in the ``Straggler Model," during each iteration, one of the users is randomly excluded from the updating process. However, for the ``Straggler Model," although not all the users participate in each round of updating, all users are involved in the subspace detection process. More importantly, the data of every user gets used multiple times (and a sufficiently large number of times when the total number of iterations is large enough). Thus, as long as each user participates in the subspace detection for large enough times, the performance of the ``Straggler Model" is comparable to that of the ``Full Model."

\subsection{Large Scale Applications}

In this subsection, we examine the performance of the proposed method when a substantial number of users contribute to its construction. Specifically, we expand the number of users to $150$, and the number of training data samples each user has is randomly selected from the range of $[1, 20]$. In addition, 30 samples are generated for testing. The process of generating data and conducting evaluations is reiterated 10 times.

Similar to Section \ref{sec:sim}, we compare the performance of the proposed ``Federated Model" with two benchmarks: ``Non-Federated Model" and ``Individual Model." As explained in Section 4.1, the dimension of the dominant subspace and the number of MFPC-scores are selected using 5-fold cross-validation. This necessitates that users have an adequate number of samples, which may not be the case for some users in the ``Federated Model" and the ``Individual Model." In the ``Federated Model," certain users might possess fewer than five samples. To overcome this challenge, we exclude these users from the cross-validation process when determining the dimension of the dominant subspace and the number of MFPC-scores. It's important to note that once these dimensions are established, all users, including those with fewer than five samples, participate in the federated feature extraction and subsequent prognostic model construction.

Similarly, in the ``Individual Model," users with fewer than five samples cannot undergo 5-fold cross-validation to determine the subspace dimension and the number of MFPC-scores. According to linear algebra principles, the dimension of the subspace and the number of MFPC-scores cannot exceed the sample size (i.e., the number of rows in the signal matrix); thus, we set the subspace dimension and the number of MFPC-scores equal to the sample size. It is worth highlighting that some users might only have one sample. For these users, individual model training is unfeasible due to insufficient data. Consequently, they are not included in the second benchmark, i.e., the ``Individual Model."

\begin{figure}[htp]
\begin{center}
    \begin{subfigure}[b]{0.4\textwidth}
         \centering
         \includegraphics[width=\textwidth]{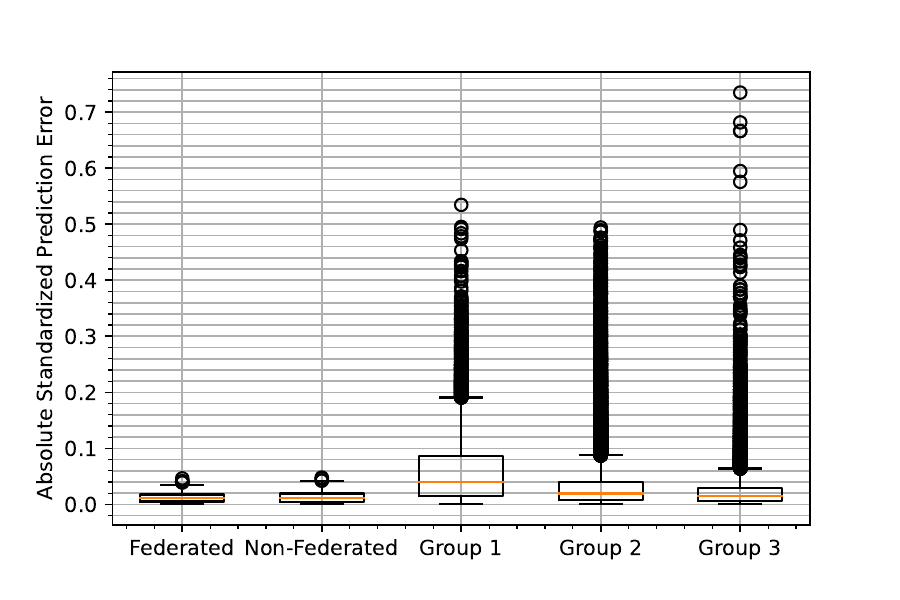}
         \caption{30\% missing values}
         \label{fig:y equals x}
     \end{subfigure}
     \hfill
     \begin{subfigure}[b]{0.4\textwidth}
         \centering
         \includegraphics[width=\textwidth]{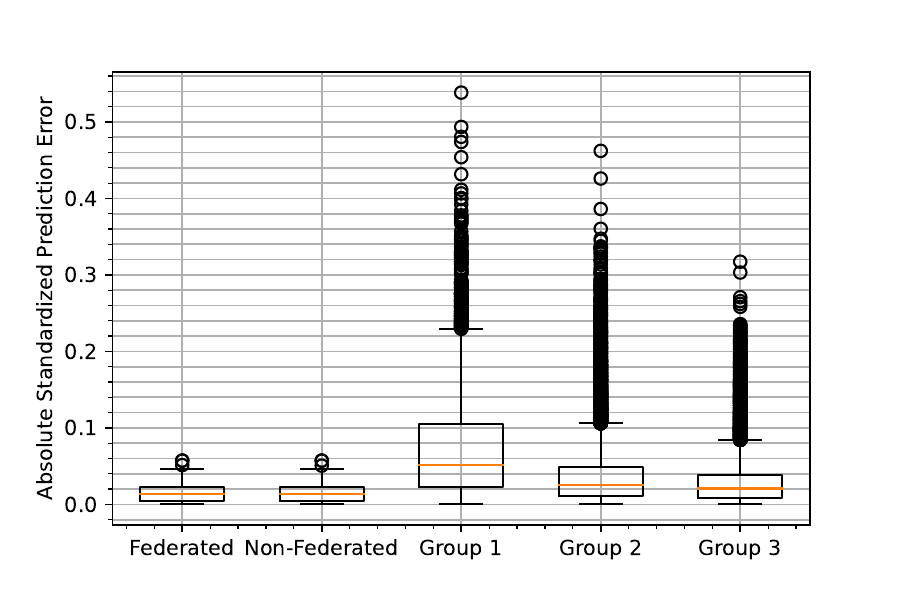}
         \caption{50\% missing values}
         \label{fig:three sin x}
     \end{subfigure}
     \hfill
     \begin{subfigure}[b]{0.4\textwidth}
          \centering
         \includegraphics[width=\textwidth]{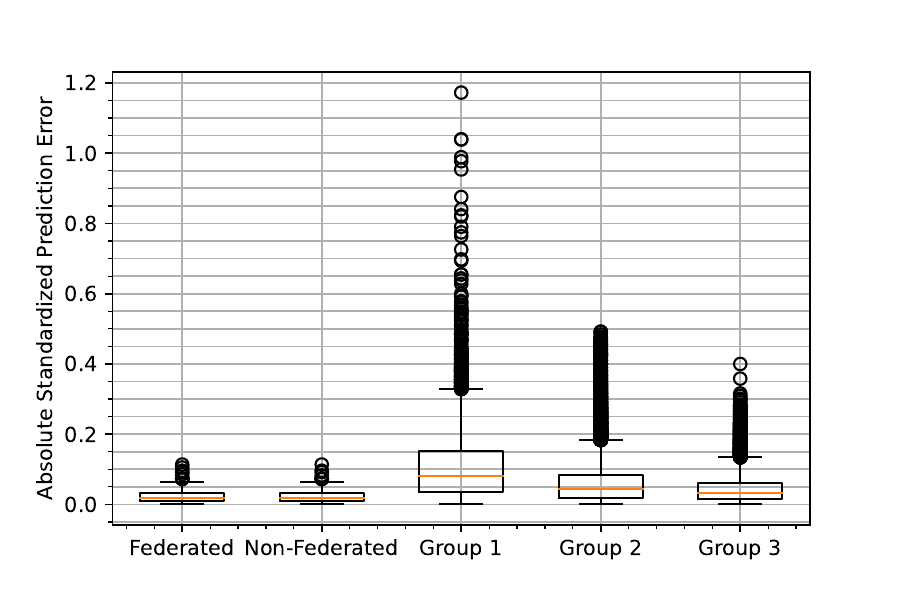}
         \caption{70\% missing values}
         \label{fig:five over x}
     \end{subfigure}
\end{center}
\caption{The prediction errors of federated model and two benchmarks\label{fig:sim:rev1}}
\vspace*{-5mm}
\end{figure}

Figure \ref{fig:sim:rev1} presents a detailed comparison of prediction errors for the federated model and the two benchmarks. Recall that we have a large number of users (i.e., $150$), which implies there are $150$ models constructed for the second benchmark (i.e., ``Individual Model"). In order to better present the prediction performance of the second benchmark, we categorize the $150$ models into three groups. Group 1 comprises users with less than 5 training samples. Group 2 includes users with 5 to 14 training samples, and Group 3 consists of users with 15 to 20 samples. As a result, there are 19, 83, and 42 users in groups 1, 2, and 3, respectively.

Similar to the observations in Section \ref{sec:sim}, Figure \ref{fig:sim:rev1} indicates that the ``Federated Model" and the ``Non-Federated Model" exhibit almost the same performance. This reaffirms that the proposed federated algorithms do not compromise the effectiveness of the prognostic method. Please notice that in Section \ref{sec:sim}, the ``Federated Model" and the ``Non-Federated Model" have the same performance. In this section, they are almost the same. This discrepancy arises from the fact that, during cross-validation in this section, users with sample sizes smaller than five are excluded from participating. In other words, the ``Federated Model" and the ``Non-Federated Model" in this section utilize slightly different datasets for cross-validation. Figure \ref{fig:sim:rev1} also illustrates that the ``Federated Model" outperforms the second benchmark, the ``Individual Model," underscoring the advantages of collaboratively constructing a federated prognostic model. Additionally, Figure \ref{fig:sim:rev1} reveals variations in performance among the three groups in the ``Individual Model." Group 1 demonstrates the poorest performance, while Group 3 exhibits the best. For instance, with a 30\% data absence, the median prediction errors (and IQR) are $0.039 (0.07), 0.019 (0.032)$, and $0.015 (0.023)$ for Groups 1, 2, and 3, respectively. This pattern persists at 50\% and 70\% data missing rates. These findings underscore the significance of having an adequate training sample size in industrial prognostics, emphasizing the importance for users, particularly those with a limited number of training samples, to participate in constructing a federated prognostic model.

\subsection{Computational Time}

In this subsection, we explore the computational time of the proposed federated prognostic model and the benchmarks. Specifically, we generate data following the settings outlined in Section \ref{sec:sim} for eight different numbers of users: $50, 100, 150, 300, 500, 800$, and $1000$, where each user has $20$ training samples. Our analysis concentrates on comparing the computational time of the ``Federated Model" and the ``Non-Federated Model." 

Recall that the proposed federated method follows a sequential approach, with Algorithm 1 executed first, followed by Algorithm 2. Algorithm 1 is iterative, and in each iteration, each user uses its local data to update the dominant subspace sequentially. Thus, the overall computational time of Algorithm 1 consists of two main components: \textit{local computation time} and \textit{communication time}. Here, \textit{local computation time} refers to the time each user spends utilizing its local data to update the dominant subspace, while \textit{communication time} denotes the time a user spends sending/receiving updates from other users. Unlike Algorithm 1, Algorithm 2 is not iterative but just requires a single round of computation. Its total computational time has three components: \textit{local computation time}, \textit{communication time}, and \textit{server computation time}. Since Algorithm 2 just involves a single round computation, the \textit{communication time} between user and server is negligible. The majority of computational time of Algorithm 2 is the \textit{server computation time}, which involves performing SVD. 

\begin{figure}[htp]
\begin{center}
    \begin{subfigure}[b]{0.4\textwidth}
         \centering
         \includegraphics[width=\textwidth]{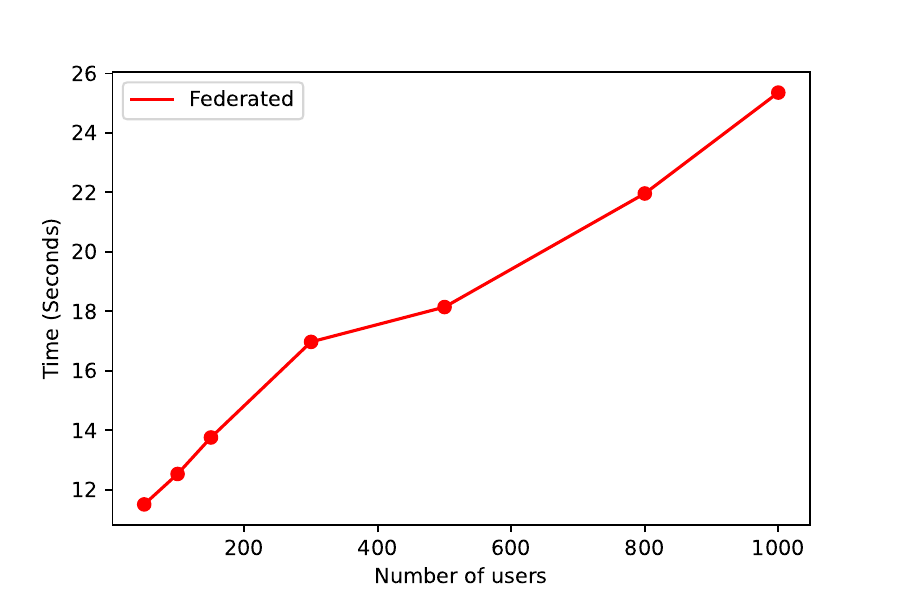}
         \caption{Federated Model Without Communication Time}
         \label{fig:y equals x}
     \end{subfigure}
     \hfill
     \begin{subfigure}[b]{0.4\textwidth}
         \centering
         \includegraphics[width=\textwidth]{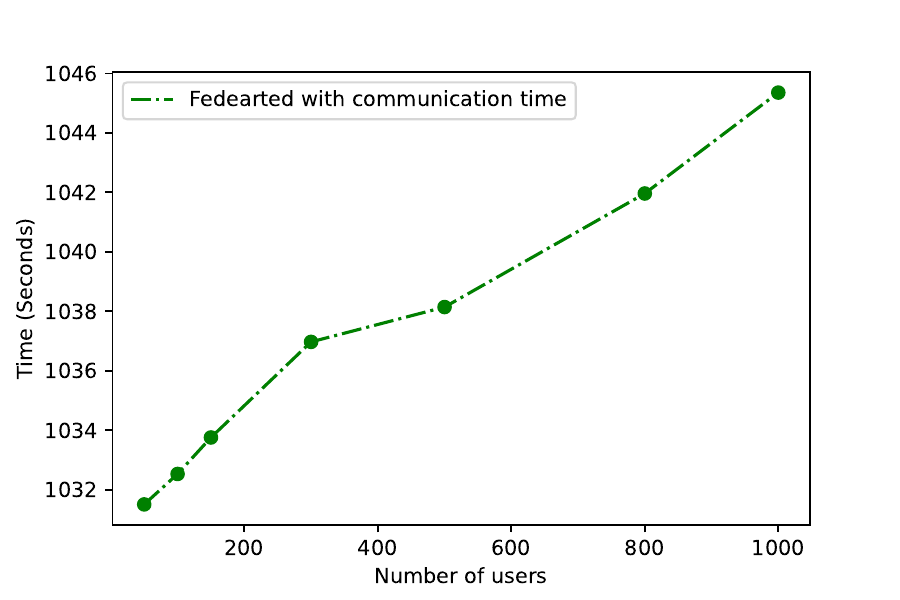}
         \caption{Federated Model with Communication Time}
         \label{fig:three sin x}
     \end{subfigure}
     \hfill
     \begin{subfigure}[b]{0.4\textwidth}
          \centering
         \includegraphics[width=\textwidth]{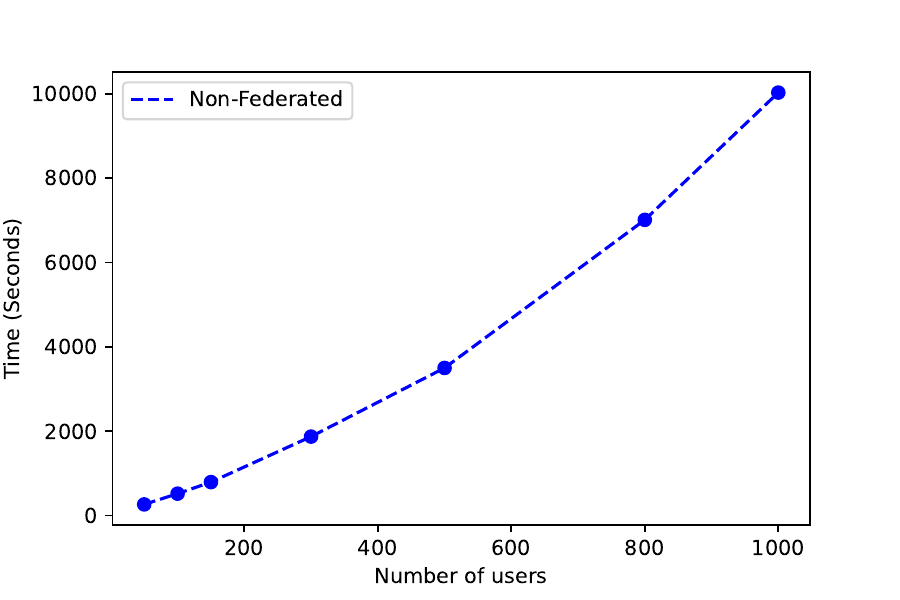}
         \caption{Non-Federated Model}
         \label{fig:five over x}
     \end{subfigure}
\end{center}
\caption{Computational time of federated and non-federated models\label{fig:sim:rev2both}}
\vspace*{-7mm}
\end{figure}

In this simulation study, we are not able to evaluate the \textit{communication time} of Algorithm 1 due to hardware constraints. This is because we have a large number of users, and each user is supposed to have a standalone computer. This implies that the total number of computers needed becomes excessively large, surpassing our computational capabilities. To address this challenge, the simulations in this section are run on a single computer, which is equipped with an Intel Core i5-1135G7 CPU with a base frequency of 2.42 GHz. \textit{The consequence of doing so is that the \textit{communication time} in Algorithm 1 is not included in the total computational time reported in Figure \ref{fig:sim:rev2both}. While we cannot provide the \textit{communication time} for Algorithm 1, we can perform a thorough analysis of it.} Specifically, in each iteration of Algorithm 1, each user utilizes its data to update the results from the former user sequentially. Therefore, the total \textit{communication time} (denoted as $t_c$) is a function of the number of users (denoted as $I$), the number of iterations (denoted as $n$), and the average \textit{communication time} between every two users (denoted as $\tau$), i.e., $t=I\times n\times \tau$. We will provide some numerical analysis on $t$ later.

Figure \ref{fig:sim:rev2both} (a) displays the computational time of the federated model without considering the \textit{communication time}. Figure \ref{fig:sim:rev2both} (b) illustrates the computational time of the federated model with the \textit{communication time}, assuming an average communication time of $\tau=2$ milliseconds between every two users. Figure \ref{fig:sim:rev2both} (c) reports the computational time of the non-federated model.

Comparing Figures \ref{fig:sim:rev2both} (a) and (c), it can be seen that when \textit{communication time} is not taken into account, the computational time of the non-federated model is much higher than that of the federated one. For instance, when there are 1000 users, the computational time for the federated model and non-federated model is approximately $25$ seconds and $10000$ seconds, respectively. Moreover, the computational time of the non-federated model experiences a rapid increase as the number of users grows, whereas the computational time of the federated model shows a more gradual increase with the rise in user numbers (excluding \textit{communication time}). For instance, the computational time for the non-federated model is $2000$ seconds with $300$ users, escalating to $10000$ seconds with $1000$ users. In contrast, the computational time of the federated model (excluding \textit{communication time}) increases modestly from $17$ seconds to $25$ seconds as the number of users rises from $300$ to $1000$. This is reasonable as the non-federated model calculates the MFPC-scores by conducting singular value decomposition (SVD) on the signal matrix containing data from all users. As the number of users increases, the size of the signal matrix grows accordingly. Consequently, the computational time required for performing SVD on this matrix experiences a substantial increase. Unlike the non-federated model, the proposed federated Algorithm 1 is incremental, which utilizes data from a single user to update results from other users. This implies that Algorithm 1 doesn't necessitate performing SVD on a large matrix, making it computationally much faster. Also, although the proposed Algorithm 2 also conducts SVD on the weight matrix, the dimension of the weight matrix is significantly smaller than that of the signal matrix. Thus, Algorithm 2 is also computationally efficient. 

Comparing Figures \ref{fig:sim:rev2both} (b) and (c), it is evident that when \textit{communication time} is considered, the computational time of the federated model is still smaller than that of the federated one. For example, if the average \textit{communication time} between every two users is $2$ milliseconds, the total computational time of the proposed method is $1031$ seconds ($50$ users), $1033$ seconds ($100$ users), $1034$ seconds ($150$ users), $1037$ seconds ($300$ users), $1038$ seconds ($500$ users), $1042$ seconds ($800$ users), and $1046$ seconds ($1000$ users).

One crucial point to emphasize is that the communication efficiency issue in industrial prognostics may not be as critical as in other applications. This is because, similar to many other data analytics methods, industrial prognostics typically involves two stages: training and real-time monitoring (i.e., testing). The training of industrial prognostic models requires joint participation from all users, while real-time monitoring is carried out individually by each user. More importantly, the training is usually conducted offline, affording users sufficient time to collaboratively train the prognostic model before its deployment for monitoring.

\section{Case Study}
\label{sec:case}

In real-world scenarios, it is common for a single company to lack sufficient historical data to independently train a prognostic model. This limitation may arise due to the challenges of having a substantial number of machines experience failures, or it may be impractical, time-consuming, or economically unfeasible to gather degradation data from a large fleet of machines. Thus, it is beneficial for multiple companies to jointly build prognostic models using their data together. For example, in the aviation industry, major aircraft engine manufacturers like General Electric (GE), Rolls-Royce, and Pratt \& Whitney may consider jointly constructing federated prognostic models. In this section, we utilize a dataset on turbofan engine degradation sourced from the NASA data repository to validate the effectiveness of the proposed federated prognostic model. We assume the existence of multiple companies and partition the dataset into distinct folds, with each company taking ownership of a specific portion of the overall dataset.

\subsection{Data Introduction and Benchmarking Models}

 The dataset contains multi-sensor degradation signals and TTFs of commercial aircraft turbofan engines \cite{saxena2008damage}. It includes 100 training engines that were run to failure and 100 test engines that were shut down at a random time before failure. In total 21 sensors were used to monitor the health condition of each of the engines, which generated 21-channel degradation signals. Readers may refer to \cite{saxena2008damage}, \cite{fang2017scalable}, and \cite{fang2017multistream} for detailed information regarding the dataset. According to \cite{fang2017multistream}, not all 21 sensors are informative for prognostics, and the four crucial sensors suggested by the authors are Total temperature at LPT outlet, Bypass Ratio, Bleed Enthalpy, and HPT coolant bleed (senor indices 4,15,17, and 20). In this article, we will use degradation signals from these 4 sensors to validate the performance of the proposed federated prognostic model.

Similar to the simulation study in Section \ref{sec:sim}, we randomly assign the training data to three users, where Users 1, 2, and 3 have $60$, $30$, and $10$ training engines, respectively. We also compare the performance of the proposed ``Federated Model" with two benchmarks, the ``Non-Federated Model" and the ``Individual Model." The proposed model and benchmarks use the same test dataset, i.e., the $100$ test engines.  We consider three levels of data incompleteness, $30\%$, $50\%$, and $70\%$. Under each incompleteness level, 15 random permutations of data are employed to evaluate the performance of the proposed method as well as the benchmarks. The rank of the dominant subspace in the proposed method and the number of MFPC-scores in all the models are chosen using 5-fold cross-validation.

\subsection{Results and Analysis}

\begin{figure}
\begin{center}
    \begin{subfigure}[b]{0.4\textwidth}
         \centering
         \includegraphics[width=\textwidth]{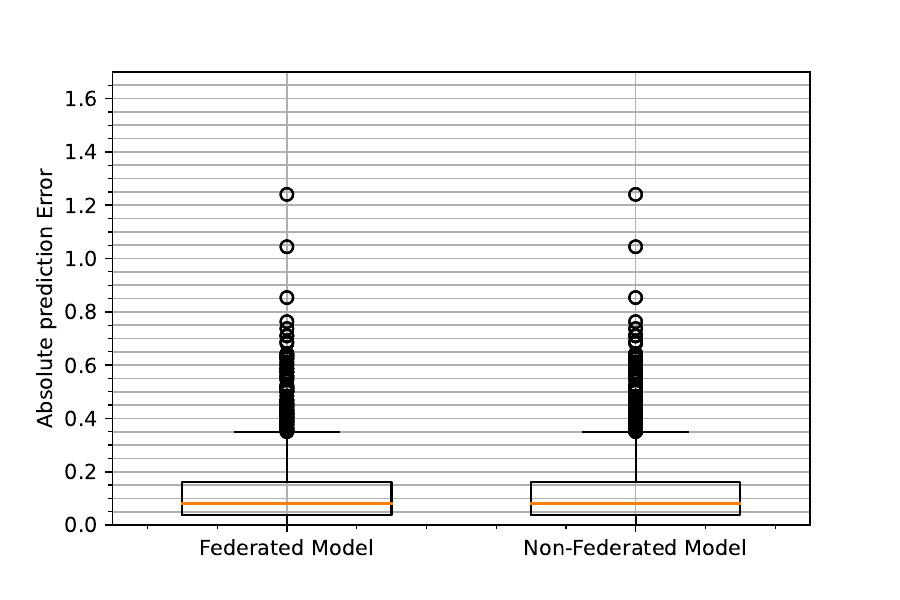}
         \caption{30\% missing values}
         \label{fig:y equals x}
     \end{subfigure}
     \hfill
     \begin{subfigure}[b]{0.4\textwidth}
         \centering
         \includegraphics[width=\textwidth]{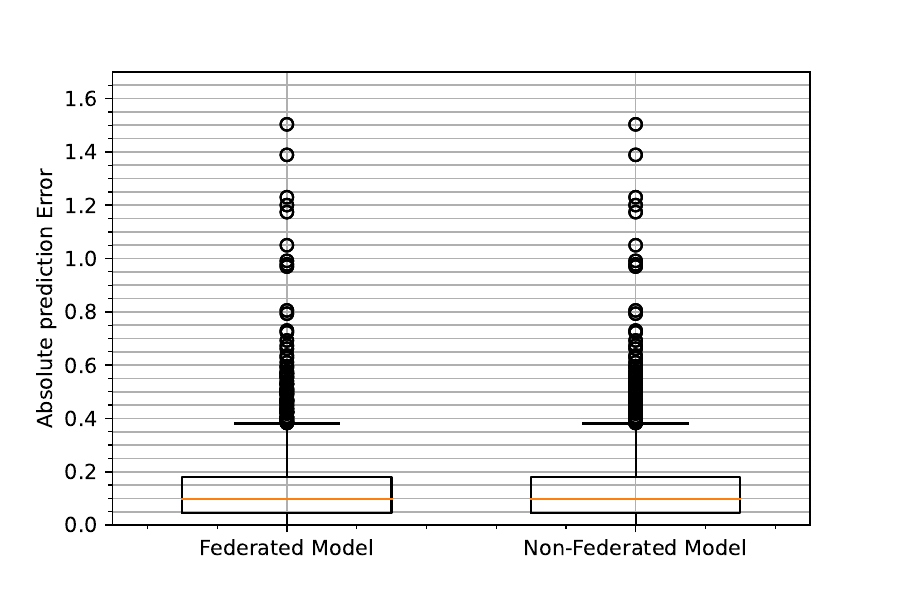}
         \caption{50\% missing values}
         \label{fig:three sin x}
     \end{subfigure}
     \hfill
     \begin{subfigure}[b]{0.4\textwidth}
          \centering
         \includegraphics[width=\textwidth]{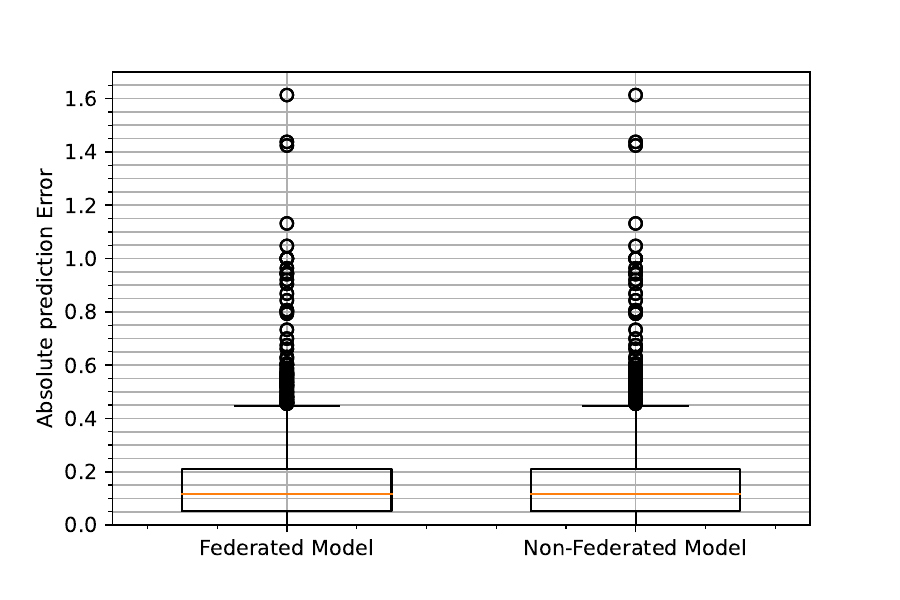}
         \caption{70\% missing values}
         \label{fig:five over x}
     \end{subfigure}
\end{center}
\vspace*{-7mm}
\caption{The prediction errors of federated model and non-federated model. \label{fig:case:federatedvsnonfederated}}
\end{figure}

The prediction performance of the proposed ``Federated Model" and the ``Non-Federated Model" is reported in Figure \ref{fig:case:federatedvsnonfederated}, which indicates that the two models achieve the same prediction accuracy and precision. For example, when the data missing level is $30\%$, the median (and IQR) of the two models are both 0.081(0.125); the median (and IQR) of them are 0.096(0.135) when the data missing level changes to $50\%$ and 0.117(0.157) when $70\%$ observations are missing. This suggests that the proposed ``Federated Model" can accomplish the same prediction accuracy and precision as the ``Non-Federated Model" which works on an aggregated dataset. 

\begin{figure}
\begin{center}
    \begin{subfigure}[b]{0.4\textwidth}
         \centering
         \includegraphics[width=\textwidth]{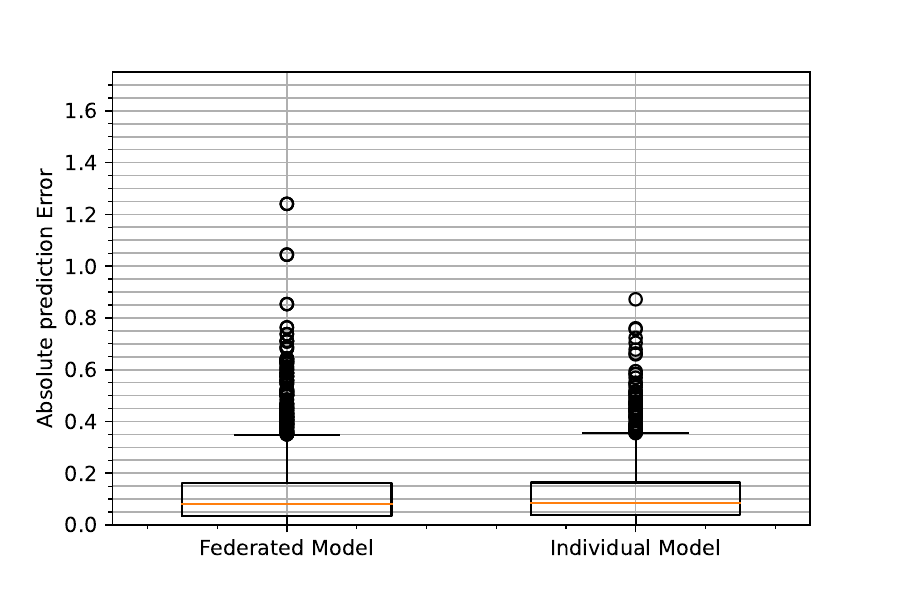}
         \caption{30\% missing values}
         \label{fig:y equals x}
     \end{subfigure}
     \hfill
     \begin{subfigure}[b]{0.4\textwidth}
         \centering
         \includegraphics[width=\textwidth]{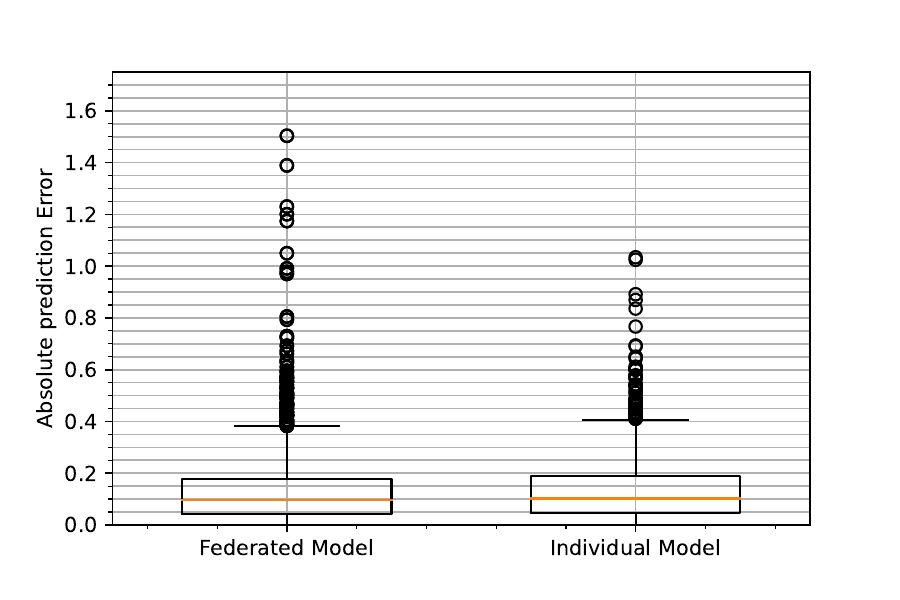}
         \caption{50\% missing values}
         \label{fig:three sin x}
     \end{subfigure}
     \hfill
     \begin{subfigure}[b]{0.4\textwidth}
         \centering
         \includegraphics[width=\textwidth]{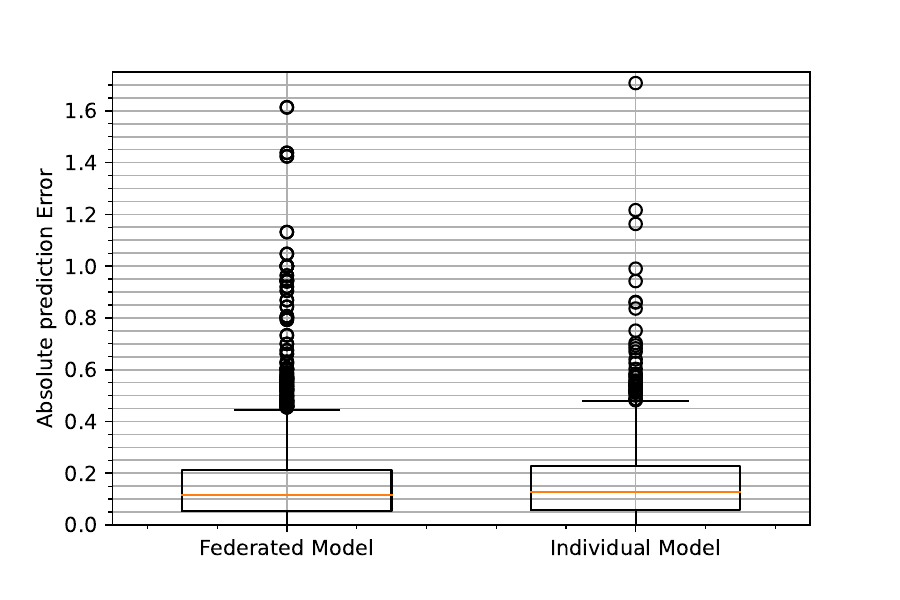}
         \caption{70\% missing values}
         \label{fig:five over x}
     \end{subfigure}
\end{center}
\vspace*{-7mm}
\caption{The prediction errors of federated model and individual model (User 1) \label{fig:case:user1}}
\end{figure}

\begin{figure}
\begin{center}
    \begin{subfigure}[b]{0.4\textwidth}
         \centering
         \includegraphics[width=\textwidth]{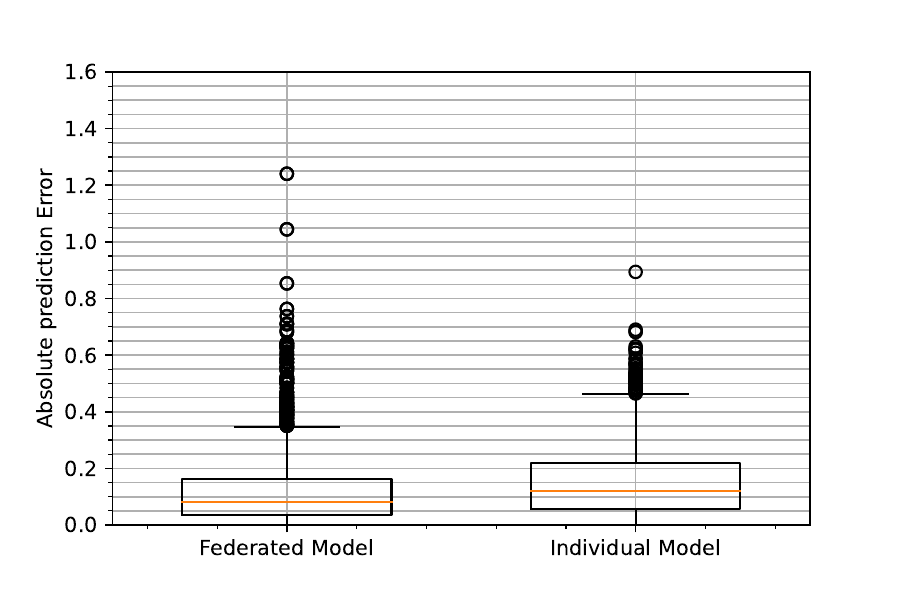}
         \caption{30\% missing values}
         \label{fig:y equals x}
     \end{subfigure}
     \hfill
     \begin{subfigure}[b]{0.4\textwidth}
         \centering
         \includegraphics[width=\textwidth]{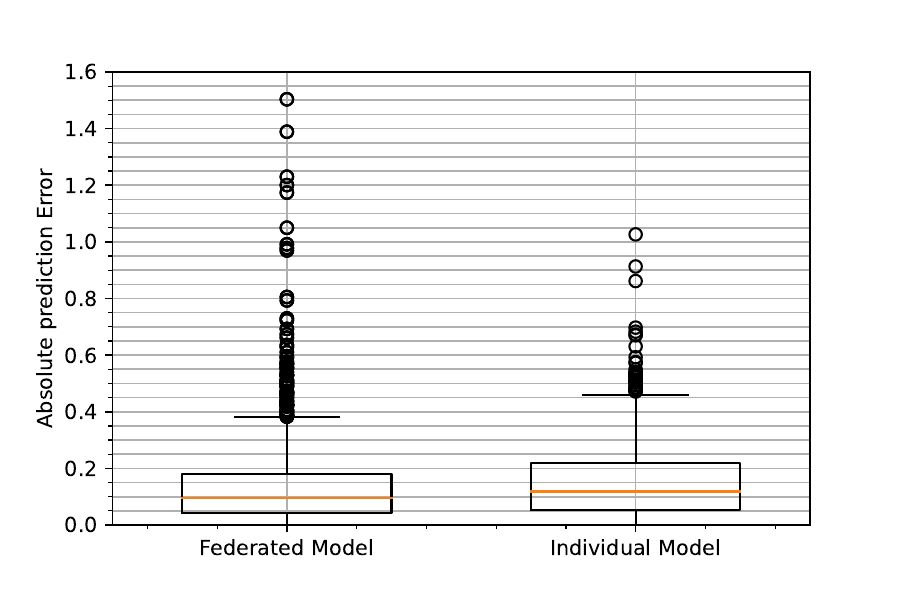}
         \caption{50\% missing values}
         \label{fig:three sin x}
     \end{subfigure}
     \hfill
     \begin{subfigure}[b]{0.4\textwidth}
         \centering
         \includegraphics[width=\textwidth]{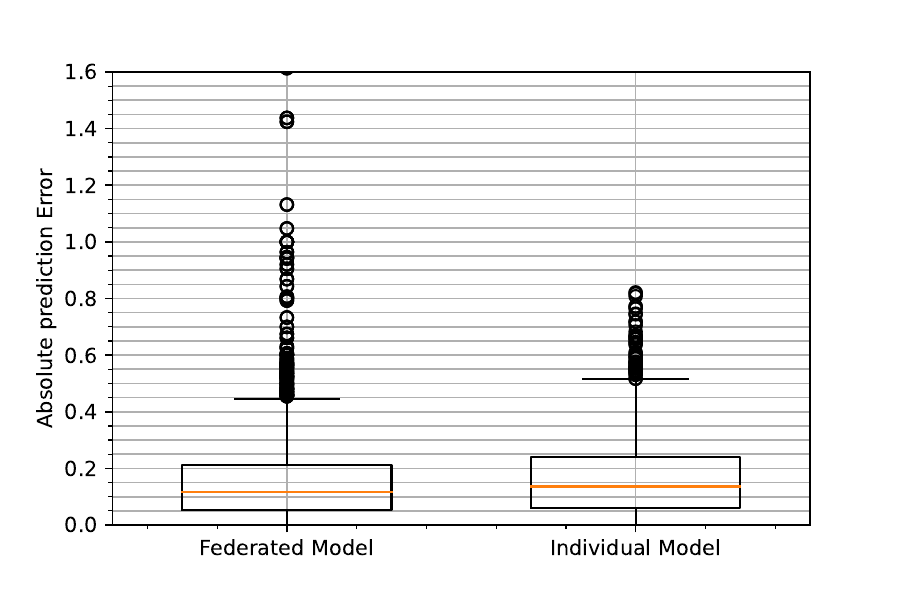}
         \caption{70\% missing values}
         \label{fig:five over x}
     \end{subfigure}
\end{center}
\vspace*{-7mm}
\caption{The prediction errors of federated model and individual model (User 2)\label{fig:case:user2}}
\end{figure}

\begin{figure}
\begin{center}
    \begin{subfigure}[b]{0.4\textwidth}
         \centering
         \includegraphics[width=\textwidth]{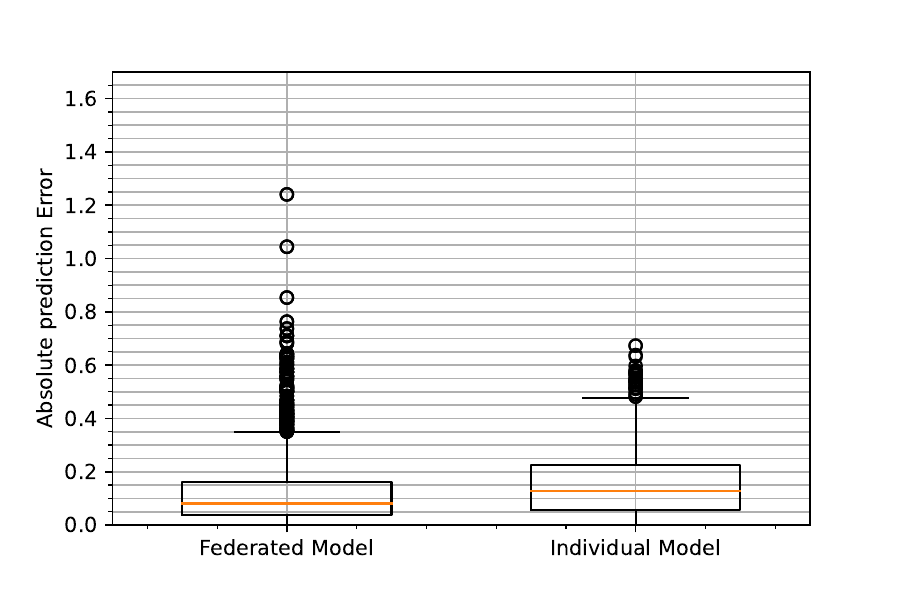}
         \caption{30\% missing values}
         \label{fig:y equals x}
     \end{subfigure}
     \hfill
     \begin{subfigure}[b]{0.4\textwidth}
         \centering
         \includegraphics[width=\textwidth]{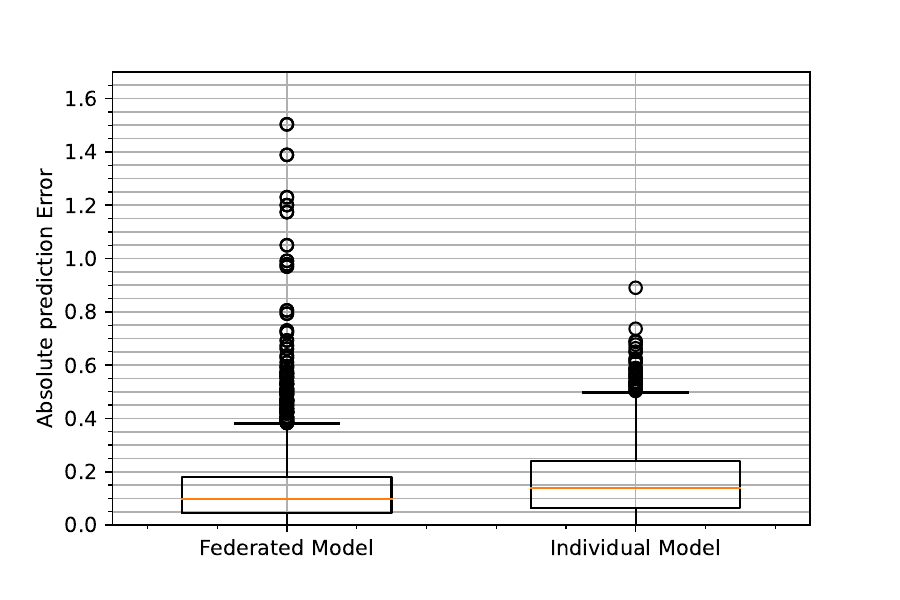}
         \caption{50\% missing values}
         \label{fig:three sin x}
     \end{subfigure}
     \hfill
     \begin{subfigure}[b]{0.4\textwidth}
         \centering
         \includegraphics[width=\textwidth]{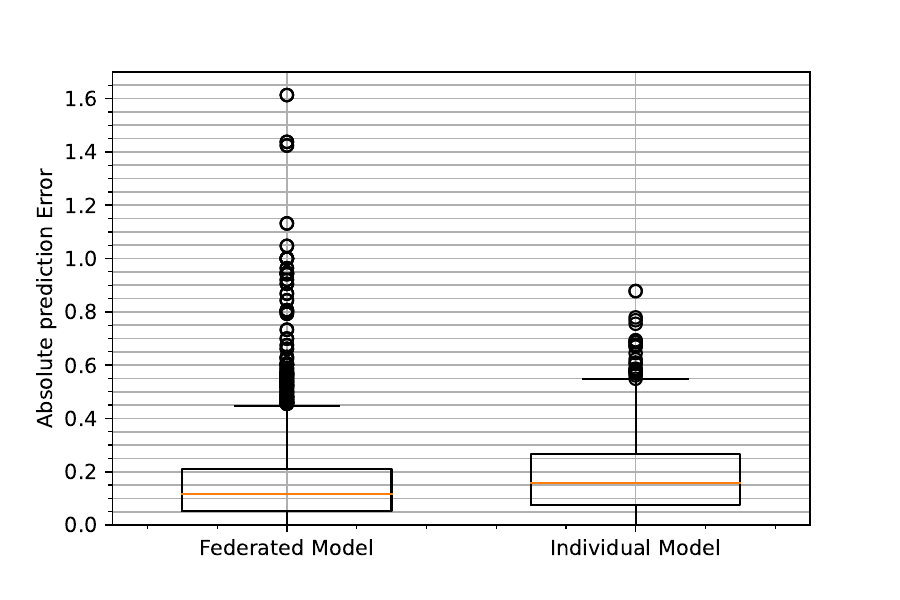}
         \caption{70\% missing values}
         \label{fig:five over x}
     \end{subfigure}
\end{center}
\vspace*{-7mm}
\caption{The prediction errors of federated model and individual model (User 3)\label{fig:case:user3}}
\end{figure}

The performance of the proposed ``Federated Model" and the second baseline model, ``Individual Model," are reported in Figures \ref{fig:case:user1}, \ref{fig:case:user2}, and \ref{fig:case:user3}, which suggest that the conclusions drawn from the simulation study still hold. The first conclusion is that the proposed ``Federated Model" works better than the ``Individual Model" at all levels of data incompleteness. For example, the medians (and IQRs) of the federated model and the individual model constructed by User 1 are 0.081(0.125) and 0.086(0.127), 0.096(0.135) and 0.102(0.144), as well as 0.117(0.157) and 0.126(0.169), when the data incompleteness levels are $30\%$, $50\%$, and $70\%$, respectively;  the medians (and IQRs) of the federated model and the individual model constructed by User 2 are 0.081(0.125) and 0.121(0.162), 0.096(0.135) and 0.118(0.166), as well as 0.117(0.157) and 0.136(0.183) when $30\%$, $50\%$, and $70\%$ data are missing, respectively. We believe this is reasonable since the federated model uses training data from all three users, and thus the size of its samples for model training is larger than that of each user. Specifically, each of the three users has a small number of samples for model training--that is--$60$ for User 1, $30$ for User 2, and $10$ for User 3, so the federated model has $100$ samples in total for model establishment. As a result, the trained federated model is more accurate than the individual models trained by each user itself. 

Another conclusion is that the performance of the individual models is affected by the size of the training samples. For example, when the data missing level is $30\%$, the medians (and IQRs) of the models constructed by Users 1, 2, and 3 are 0.086(0.127), 0.121(0.162), and 0.129(0.169), respectively. Similarly, when $50\%$ data are missing, the medians (and IQRs) are respectively 0.102(0.144), 0.118(0.166), and 0.140(0.176); when the data incompleteness level is $70\%$, they are respectively 0.126(0.169), 0.136(0.183), and 0.159(.188). This again proves the importance of sample size in establishing a high-performance prognostic model and thus verifies the value and necessity of federated learning-based prognostic when the size of samples owned by each individual user is small.

\section{Conclusions}\label{sec:conc}
In this article, we proposed a federated prognostic model that allows multiple users to collaboratively train the model while keeping the data of each user local and confidential. The proposed model works for applications with multiple stream degradation signals, which may be incomplete (i.e., some observations are missing). As some of the existing works in remaining useful lifetime prediction, the prognostic model used in this paper first uses multivariate functional principal component analysis (MFPCA) to fuse the multi-stream degradation signals; the fused features (known as MFPC-scores) coupled with the times-to-failure (TTFs) are utilized to build a (log)-location-scale regression model for failure time prediction. To address the challenges that 1) some observations of degradation signals are missing and thus classic algorithms for MFPC-score computation cannot be used, and 2) the feature extraction (i.e., MFPC-score computation) should be federated, we proposed a federated feature extraction method that enables multiple users to utilize their incomplete degradation signals to compute their MFPC-scores while keeping each user's data local and confidential. 

The proposed federated feature extraction method consists of two steps. First, we developed a federated algorithm to detect the dominant subspace of the degradation signal matrix, which provides a general set of basis vectors of the dominant subspace. Next, we propose a federated algorithm that uses the general set of basis vectors of the dominant subspace to calculate the dominant eigenvectors as well as the MFPC-scores. Simulated datasets as well as a degradation dataset from aircraft engines were used to evaluate the performance of the proposed federated prognostic model. The results indicated that the proposed federated prognostic model accomplished the same accuracy and precision for failure time prediction as that of the non-federated model using an aggregated dataset from all users. This suggests that the proposed model can protect the data privacy of users while not compromise the performance of failure time prediction. In addition, the results showed that the performance of the proposed federated model is better than that of the prognostic models constructed by each user itself, especially when a user has a small number of data samples for model training. This further verifies the value and necessity of the developed federated prognostic model.

\newpage

\section{Appendix}

\subsubsection{Proof of Proposition 1}

It is known that $\boldsymbol{U}_{dominant}$ and $\boldsymbol{P}$ represent the same set of eigenvectors defined in two different coordinate systems. $\boldsymbol{P}\in\mathbb{R}^{K\times K}$ is defined in the \textit{new coordinate system} constructed by the columns of $\boldsymbol{U}_{new}\in\mathbb{R}^{N\times K}$. In the meanwhile, $\boldsymbol{U}_{dominant}\in\mathbb{R}^{N\times K}$ and $\boldsymbol{U}_{new}$ itself are defined in the \textit{default coordinate system} (the same coordinate system in which the signal matrix $\boldsymbol{X}$ is defined). Since $\boldsymbol{U}_{dominant}$ is a set of vectors defined in the \textit{default coordinate system}, $\boldsymbol{U}_{new}^\top\boldsymbol{U}_{dominant}$ gives its representation in the \textit{new coordinate system}, which is known as $\boldsymbol{P}$. In other words, $\boldsymbol{P}=\boldsymbol{U}_{new}^\top\boldsymbol{U}_{dominant}$ or $\boldsymbol{U}_{new}\boldsymbol{P}=\boldsymbol{U}_{dominant}$. MFPC-scores $\boldsymbol{Z}=\boldsymbol{U}_{dominant}^\top\boldsymbol{\tilde{X}}=(\boldsymbol{U}_{new}\boldsymbol{P})^\top\boldsymbol{\tilde{X}}=\boldsymbol{P}^\top\boldsymbol{U}_{new}^\top\boldsymbol{\tilde{X}}=\boldsymbol{P}^\top\boldsymbol{U}_{new}^\top(\boldsymbol{X}-\boldsymbol{\bar{X}})=\boldsymbol{P}^\top\boldsymbol{U}_{new}^\top\boldsymbol{X}-\boldsymbol{P}^\top\boldsymbol{U}_{new}^\top\boldsymbol{\bar{X}}=\boldsymbol{P}^\top\boldsymbol{W}-\boldsymbol{P}^\top\boldsymbol{\bar{W}}=\boldsymbol{P}^\top(\boldsymbol{W}-\boldsymbol{\bar{W}})=\boldsymbol{P}^\top\boldsymbol{\tilde{W}}$.








\subsubsection{Proof of Equation \eqref{eq5} \cite{fang2017multistream}}

Recall that $ {y}_{ij} = \gamma_0 +\int_0^T \boldsymbol{\gamma}(t)^\top \boldsymbol{x}_{ij}(t)dt+\sigma\epsilon_{ij}$, where $\boldsymbol{x}_{ij}(t) = \boldsymbol{\mu}(t) + \sum_{k=1}^\infty  \boldsymbol{\zeta}_{ijk}\boldsymbol{\Psi}_k(t)$ and $\boldsymbol{\gamma}(t)=\sum_{k=1}^\infty \beta_k\boldsymbol{\Psi}_k(t)$. Also, it is known that $\int_0^T\boldsymbol{\Psi}_m(t)\boldsymbol{\Psi}_n(t)dt=0$ if $m\neq n$ and $\int_0^T\boldsymbol{\Psi}_m(t)\boldsymbol{\Psi}_n(t)dt=1$ if $m=n$. Therefore, we have the following:

\begin{align*}
{y}_{ij} & =\gamma_0 +\int_0^T \boldsymbol{\gamma}(t)^\top \boldsymbol{x}_{ij}(t)dt +\sigma\epsilon_{ij}\\
 & = \gamma_0 +\int_0^T \boldsymbol{\gamma}(t)^\top \boldsymbol{\mu}(t)dt +\int_0^T \boldsymbol{\gamma}(t)^\top \left( \sum_{k=1}^\infty  \boldsymbol{\zeta}_{ijk}\boldsymbol{\Psi}_k(t)dt\right)dt+\sigma\epsilon_{ij} \\
 & = \gamma_0 +\int_0^T \boldsymbol{\gamma}(t)^\top \boldsymbol{\mu}(t)dt +\int_0^T \left(\sum_{k=1}^\infty \beta_k\boldsymbol{\Psi}_k(t) \right)\left(  \sum_{k=1}^\infty  \boldsymbol{\zeta}_{ijk}\boldsymbol{\Psi}_k(t)dt\right) +\sigma\epsilon_{ij}\\
 & =\beta_0 +  \sum_{k=1}^\infty  \boldsymbol{\beta}_{k}\boldsymbol{\zeta}_{ijk}+\sigma\epsilon_{ij}\\
 & \approx \beta_0 +  \sum_{k=1}^K \boldsymbol{\beta}_{k}\boldsymbol{\zeta}_{ijk}+\sigma\epsilon_{ij}
\end{align*}

\noindent where $\beta_0= \gamma_0 +\int_0^T \boldsymbol{\gamma}(t)^\top \boldsymbol{\mu}(t)dt$.

\section{Data Availability Statement}

The data that support the findings of this study are openly available in NASA Prognostics Center of Excellence Data Set Repository at https://www.nasa.gov/content/prognostics-center-of-excellence-data-set-repository.

\bibliographystyle{unsrt}
\bibliography{Paper}



\end{document}